\documentclass[letterpaper,twocolumn]{article}

\usepackage[utf8]{inputenc}
\usepackage[T1]{fontenc}
\usepackage{times}
\makeatletter
\DeclareSymbolFont{largesymbols}{OMX}{cmex}{m}{n}
\makeatother

\usepackage{graphicx}
\usepackage{amsmath,amssymb,amsthm}
\usepackage{booktabs}
\usepackage{float}
\usepackage{makecell}
\usepackage{comment}
\usepackage{microtype}
\usepackage[hyphens]{url}

\usepackage[numbers,sort&compress]{natbib}
\makeatletter
\def\addcontentsline#1#2#3{}

\newlength\titlebox 
\def\maketitle{\par
 \begingroup
   \def\thefootnote{\fnsymbol{footnote}}
   \def\@makefnmark{\hbox to 0pt{$^{\@thefnmark}$\hss}}
   \twocolumn[\@maketitle] \@thanks
 \endgroup
 \setcounter{footnote}{0}
 \let\maketitle\relax \let\@maketitle\relax
 \gdef\@thanks{}\gdef\@author{}\gdef\@title{}\let\thanks\relax}
\def\@maketitle{\vbox to \titlebox{\hsize\textwidth
 \linewidth\hsize \vskip 0.35in minus 0.125in \centering
 {\LARGE\bf \@title \par} \vskip 0.2in plus 1fil minus 0.1in
 {\def\and{\unskip\enspace{\rm and}\enspace}%
  \hbox to \linewidth\bgroup\normalsize \hfil\hfil
    \hbox to 0pt\bgroup\hss \begin{tabular}[t]{c}\Large\@author
                            \end{tabular}\hss\egroup
    \hfil\hfil\egroup}
  \vskip 0.5in plus 2fil minus 0.1in
}}
\renewenvironment{abstract}{\centerline{\bf
 Abstract}\vspace{0.5ex}\begin{quote}\small}{\par\end{quote}\vskip 1ex}

\renewcommand\section{\@startsection{section}{1}{\z@}%
   {-3.0ex plus -0.5ex minus -.2ex}{6pt plus 2pt minus 1pt}%
   {\Large\bf\centering}}
\renewcommand\subsection{\@startsection{subsection}{2}{\z@}%
   {-2.5ex plus -0.5ex minus -.2ex}{5pt plus 2pt minus 1pt}%
   {\large\bf\raggedright}}
\renewcommand\subsubsection{\@startsection{subsubsection}{3}{\z@}%
   {-2.5ex plus -0.5ex minus 0.0ex}{4pt plus 2pt minus 1pt}%
   {\normalsize\bf\raggedright}}
\skip\footins 9pt plus 4pt minus 2pt
\def\footnoterule{\kern-3pt \hrule width 5pc \kern 2.6pt }
\def\@normalsize{\@setsize\normalsize{12pt}\xpt\@xpt}
\def\small{\@setsize\small{11pt}\ixpt\@ixpt}
\def\footnotesize{\@setsize\footnotesize{11pt}\ixpt\@ixpt}
\def\scriptsize{\@setsize\scriptsize{9pt}\viipt\@viipt}
\def\tiny{\@setsize\tiny{7pt}\vipt\@vipt}
\def\large{\@setsize\large{13pt}\xipt\@xipt}
\def\Large{\@setsize\Large{15pt}\xiipt\@xiipt}
\def\LARGE{\@setsize\LARGE{17pt}\xivpt\@xivpt}
\makeatother

\theoremstyle{plain}
\newtheorem{theorem}{Theorem}
\newtheorem{lemma}{Lemma}

\theoremstyle{definition}
\newtheorem{definition}{Definition}
\theoremstyle{remark}

\usepackage{hyperref}
\hypersetup{
  colorlinks=true, linkcolor=blue, citecolor=blue, urlcolor=blue,
  breaklinks=true,
  pdftitle={Emergent Models: Intelligence from Tiny Substrates},
  pdfauthor={Giacomo Bocchese, Nicola Giacobbo, Etienne Guichard, James Wiles, Akshaj Devireddy},
  pdfsubject={cs.NE},
  pdfkeywords={emergent models, cellular automata, evolutionary computation,
               latent universality, extrapolation}}
\usepackage{cleveref}
\title{Emergent Models: Intelligence from Tiny Substrates}
\author{
    Giacomo Bocchese$^{1,2}$,
    Nicola Giacobbo$^{2}$,
    Etienne Guichard$^{3,2}$,
    James Wiles$^{1}$,
    Akshaj Devireddy$^{1,2}$ \\
    \mbox{}\\
    $^{1}$Wolfram Institute,
    $^{2}$Emergent Computing,
    $^{3}$\O stfold University College
}
\begin{document}
\maketitle
\begin{abstract}
Emergent Models (EMs) are a machine learning paradigm based on simple yet open-ended substrates, such as cellular automata, in which modeling is treated not as the learning of a closed-form input-output map but as the emergence, within simple dynamical systems, of computational behaviors that solve external tasks. Such substrates typically iterate a fixed local rule over a latent space for an adaptive number of steps, with an interface linking the latent state to external input/output signals. Training proceeds by evolutionary search. We hypothesize that some instances of this framework are biased toward global generalization: capturing the rule generating the data over its full domain, and therefore extrapolating beyond the training range. Theoretically, we prove that some EMs are \emph{latent-universal}: with the update rule and interface held fixed, they can realize any partial computable function by varying only the initial condition of the latent state. Empirically, we study a zoo of minimal EM instantiations across discrete and continuous substrates, showing that local-recursive computation at a tiny scale (tens to hundreds of parameters) can extrapolate exactly on simple arithmetic functions, can support control behaviour and online adaptation, while still exposing several limitations. This work is foundational: it does not propose a competitive architecture, but a framework meant to widen the design space of machine learning beyond differentiable feed-forward maps.
\end{abstract}

\vspace{0.1em}
\noindent\footnotesize\textit{Code \& demo:} \href{https://emergentcomputing.github.io/em-paper/}{https://emergentcomputing.github.io/em-paper/}
\normalsize
\vspace{0.3em}

\section{Introduction}
Sutton's "Bitter Lesson" argues that the strongest long-run progress in AI comes from general methods that leverage compute, rather than from direct top-down engineering \citep{sutton2019bitter}. A similar intuition appears in artificial life and open-ended evolution, where complex life-like behaviors are studied as emerging from simple yet general physical substrates under long-run evolutionary processes \citep{dolson2024oee}. This motivates the search for substrates governed by simple rules that can be emulated on a computer, in which external evolutionary pressure can select for the emergence of task-solving behaviors. Emergent Models (EMs) formalize this perspective by turning such dynamical systems into machine learning models, coupled to external inputs and outputs through an encoding and decoding interface.

But which notion of openness is relevant for machine learning? And is there a precondition for a substrate to "leverage compute"? We argue that both questions have the same answer: expressivity, the set of functions a model can theoretically represent. Search can only select among functions a model can express: a target behavior outside this range is unreachable at any compute budget.

Because the term expressivity is often used informally and varies across contexts \citep{butoi2024computational,wiki_expressive_power_cs,guhring2020}, we define explicitly the notion relevant here: \emph{global expressivity}, the class \(G_\phi\) of functions that a family of models \(\phi_\theta\), parametrized by \(\theta\), can represent exactly or approximate to arbitrary precision \emph{over their full, possibly open domain}. The global qualifier reflects the goal of modeling, which is to capture the rule generating the data rather than to fit its values on a bounded region: a model that internalizes the rule can extrapolate by construction, while a local fit does not \citep{kumar2023,marcus2018,chollet2019}.

Much of traditional deep learning is based on feed-forward neural networks (FFNNs) \citep{goodfellow2016deep}. These fixed-depth, closed-form mappings typically extrapolate poorly beyond their training range \citep{Haley1992ExtrapolationLO}, reflecting limitations in both expressivity and inductive bias\footnote{\begingroup\scriptsize\raggedright
Expressivity concerns whether a target function lies in the representable class at all (an existence question), while inductive bias concerns which representable solution training tends to select from finite data. The two are related: a model can only be biased toward a solution that it is first able to express, so expressivity is a precondition for inductive bias.
\par\endgroup}\citep{baxter2011model}. On the expressivity side, the limitation is structural: universal approximation theorems for FFNNs \citep{hornik1991} guarantee arbitrarily good approximation only over compact domains, and over open domains what a network can represent or approximate is dictated by the asymptotic behavior induced by its activation function. For scalar functions \(\mathbb{R}\to\mathbb{R}\), sigmoidal networks converge to constants and ReLU networks become asymptotically linear. As a result, FFNNs cannot globally represent functions such as \(\sin(x)\) or \(e^x\) without task-specific engineering\footnote{\begingroup\scriptsize\raggedright
For example, periodic activations or Fourier features for \(\sin(x)\).
\par\endgroup}\citep{vannuland2023noncompact}.

In traditional recurrent or autoregressive architectures, such as standard (non-reasoning) RNNs and Transformers, each datapoint prediction is obtained through a single forward pass \(x_i \mapsto y_i\), favoring direct one-shot mapping from input to output \citep{ghojogh2023rnnlstm,radford2018gpt,wolfram2023chatgpt}.
A growing body of work suggests that expressivity is increased when models are allowed to perform \emph{multiple iterative updates per input}, rather than a single forward pass. Adaptive Computation Time makes this explicit in recurrent networks by applying the same update function to the hidden state for a variable number of steps before producing an output \citep{graves2016act}. Reasoning LLMs apply a similar principle in the token space, expanding the prediction process through intermediate reasoning traces before producing an answer \citep{guo2025deepseekr1}; these models have achieved strong gains on arithmetics, coding, and ARC-AGI reasoning puzzles \citep{chollet2019,openai2024o1}, and under suitable assumptions are Turing complete \citep{jiang2025softmaxcottc}, however, they remain tied to very large neural networks and appear brittle on long-sequence extrapolation \citep{shojaee2025illusion,abbe2024globality}.

This raises a question about compression: how small can the iterated update function be while still supporting high global expressivity, and does smallness itself carry favorable inductive biases for extrapolation? Compact recursive models such as Hierarchical Reasoning Models (HRMs) and Tiny Recursive Models (TRMs) match mid-size LLMs on ARC-AGI-style reasoning with \(\approx10^3\) fewer parameters \citep{wang2025hrm,jolicoeurmartineau2025trm}. Neural GPUs and Neural Cellular Automata (NCAs) push it further by iterating a small neural update function locally across a spatial latent medium \citep{kaiser2016neuralgpu,mordvintsev2020growingnca}: a Neural GPU trained on binary multiplication up to 20-bit inputs generalizes perfectly to 2000 bits, and NCAs reach significant ARC-AGI scores with \(\approx10^5\) fewer parameters than comparably performing LLMs \citep{guichard2025arcnca,xu2025ncaarcagi}. Across these lines of work, a common pattern emerges: local, recurrent, compact neural updates appear to favor algorithmic generalization while reducing parameter size.

The emphasis on iteration is grounded in computation theory: in classical models of computation (eg. Turing machines), complex algorithmic behavior arises from repeatedly applying simple update rules to a mutable memory state, with computation time and memory allowed to vary depending on input and task complexity. Looping is not an implementation detail, but one of the core ingredients that makes general computation possible \citep{sipser2012introduction,harel2004algorithmics}. The strongest general notion of global expressivity in this algorithmic setting is computational universality, or Turing completeness \citep{sipser2012introduction,Wolfram2002}. A universal system, given unbounded memory and time, can compute any partial computable function: in discrete domains this yields exact global representation, while for continuous computable functions it permits arbitrary-precision global approximation under a proper discretization convention \citep{miller2007computability_notes,weihrauch2000computable,braverman2005computingreals}. Universality says nothing about trainability or efficiency, but it sets a necessary condition: if a model class cannot realize arbitrary computable procedures even in principle, there are hard limits on what it can express.

Cellular automata (CAs) are a natural substrate in this setting. They combine a simple local rule with iterative dynamics over a potentially unbounded medium, and some, including Conway's Game of Life and Rule 110, are Turing complete \citep{cook2004universality,rendell2016turing}. Several notions of universality are studied for CAs; here we focus on \emph{strong} universality\footnote{\begingroup\scriptsize\raggedright
Three notions of universality are most commonly distinguished for CAs. \emph{Weak} universality employs an infinite periodic pattern, \emph{Strong} universality requires an infinite quiescent background, \emph{Intrinsic} universality is stronger still: can simulate any other CA under a fixed block rescaling of space and time. \citep{ollinger2008universalities}.
\par\endgroup}, which most closely aligns with classical Turing's notion for computing functions.

The use of CAs in machine learning is not entirely new, and so far has followed two main approaches: CA reservoirs and NCAs. Reservoir methods treat the automaton as a fixed dynamical medium and train only an external linear readout of its trajectories \citep{yilmaz2015reca,nichele2017deepreca}. NCAs instead train the update function itself, typically a small convolutional neural network applied locally \citep{variengien2021cartpolenca,guichard2024critically}. In both cases the initial condition is not treated as an explicitly learnable object; instead our framework trains it directly, optionally jointly with the update rule, so that the initial state acts as a program shaping the automaton's evolution and consequently the predictions of the model. 

In this work, we formalize this perspective through a framework that turns dynamical systems into machine learning models. In its strongest ideal regime, which we call \emph{latent universality} (\cref{thm:lu}), the update rule is fixed and varying only the initial state suffices to express any computable function.

Our experiments deliberately study a simpler setting: they do not realize latent universality, and no experimental claim depends on Theorem~\ref{thm:lu}. Rather than fine-tuning a single construction toward the full theory, we examine a zoo of minimal Emergent Model instantiations to convey the idea and philosophy behind the framework, as befits the opening stage of this research. These show that local-recursive modeling at a tiny scale (20-300 parameters) can extrapolate on simple arithmetic functions and can produce control behaviors, indicating that such substrates are meaningful in principle, not that they are easy to train or competitive with mature task-specific methods.

\section{Emergent Models}

\subsection{Intuition}
A broad class of latent-space reasoning models can be summarized schematically as:
\[
x \;\to\; E(x) \;\to\; s_0 \;\xrightarrow{\,f^T\,}\; s_T \;\to\; D(s_T) \;\to\; y
\]
where a fixed update function \(f\) is iterated on a latent state \(s\) for a number of steps \(T(s_0)\), which can vary and may depend on the initial condition. In this view, the latent state provides memory ("computational space"), while repeated application of \(f\) provides computation time. The encoder \(E\) writes inputs into the latent space, and the decoder \(D\) reads outputs from it. Encoder and decoder form an \emph{interface}: a communication protocol between the latent space and the external world.

\begin{figure}[H]
    \centering
    \includegraphics[width=1\linewidth]{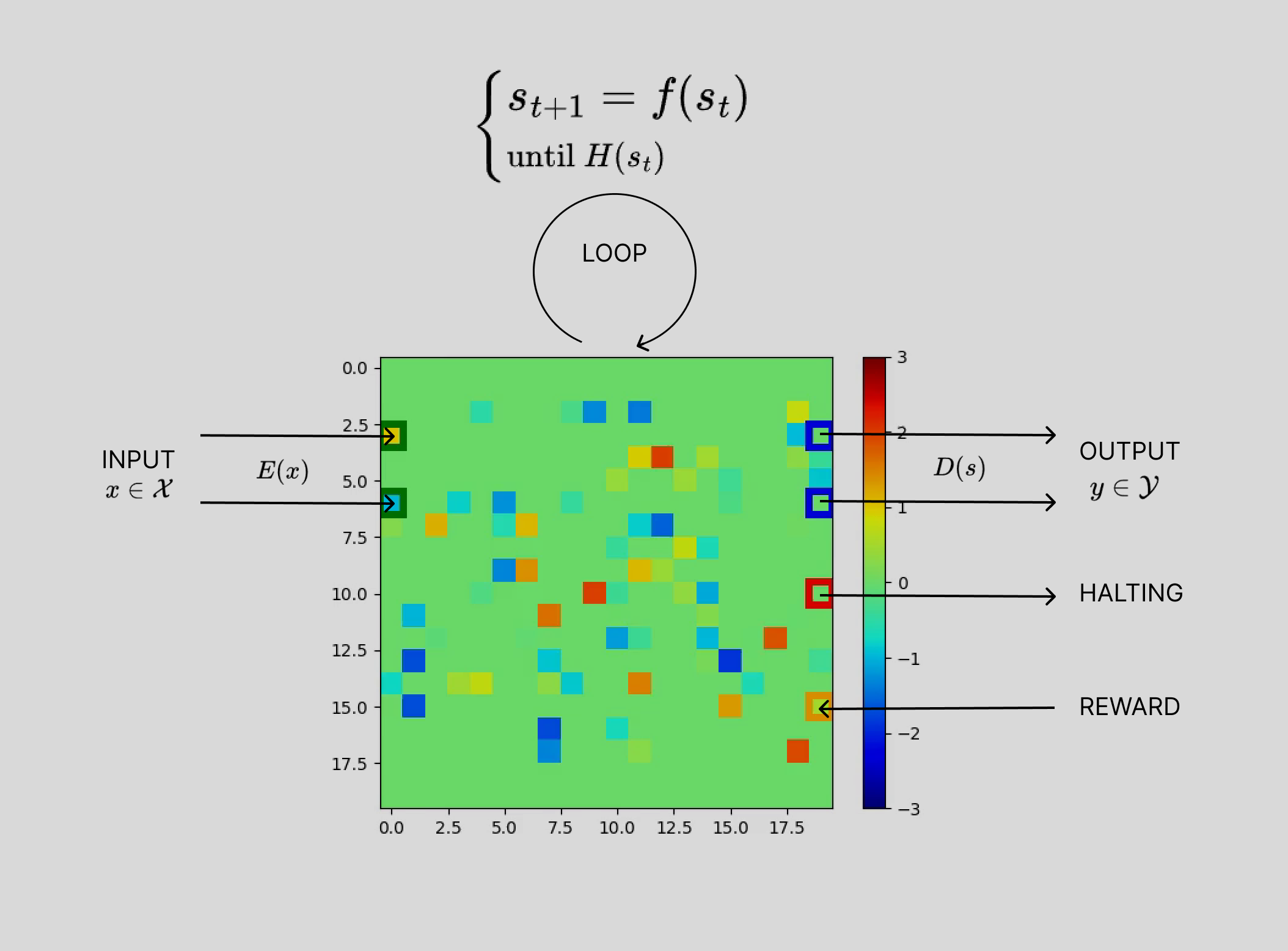}
    \caption{Emergent Model with a two-dimensional continuous-valued lattice as latent state, \(s\in\mathbb R^{20\times20}\). 
    Coloured borders mark the ports: input (green), output (blue), halting (red), and reward (orange). 
    \emph{Interface:} the input \(x\in\mathbb{R}^2\) is injected at the two input ports through the encoder \(E(x)\); at halting, the decoder \(D(s_T)\) reads the output \(y\in\mathbb{R}^2\) from the two output ports. 
    \emph{Update:} the transition rule is iterated, \(s_{t+1}=f(s_t)\), for an adaptive number of steps, halting when \(H(s_t)=1\). The role of the reward port is explained in Section~\ref{sec:retention}.}
    \label{fig:em-loop}
\end{figure}

Figure~\ref{fig:em-loop} shows a didactical instance: a continuous two-dimensional lattice with an interface based on fixed cell positions: inputs and outputs are written and read at designated locations (ports).

\begin{figure}[H]
    \centering
    \includegraphics[width=1\linewidth]{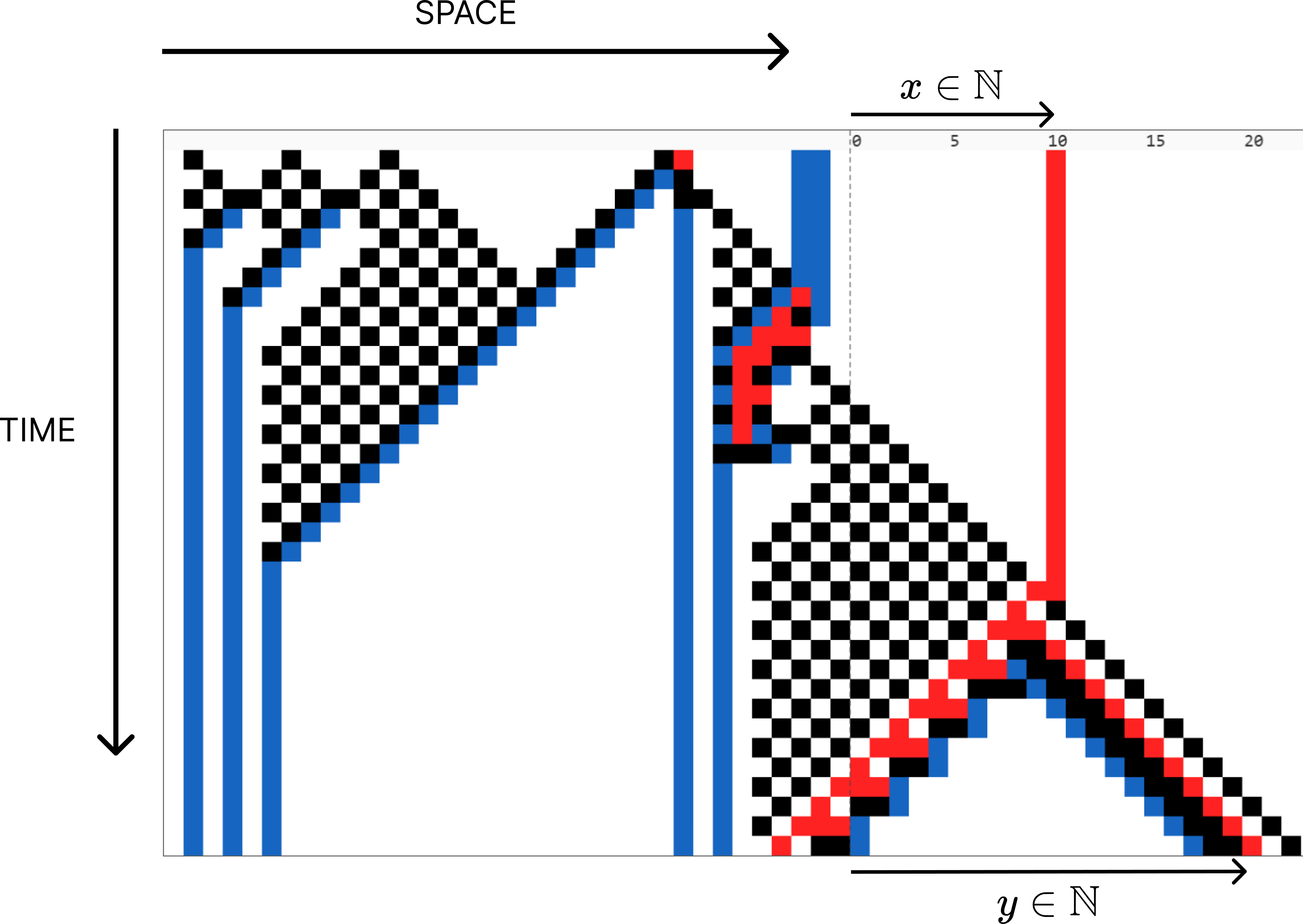}
    \caption{Diagram representing the latent state of EM43, a discrete Emergent Model, evolving in time. It is based on a four-valued, radius-1 cellular automaton over a one-dimensional unbounded tape, with state \(s\in\{0,1,2,3\}^{\mathbb N}\). This uses a different interface from~\cref{fig:em-loop}: input is not written to a fixed port, but placed at a position that depends on its value. The same holds for the output.
    \emph{Encoding:} an integer input \(x\in\mathbb N\) is written as a red cell at position \(x_0+x\), i.e.\ at offset \(x\) from a fixed origin \(x_0\). In this case input \(x=10\).
    \emph{Decoding:} at halting, the output \(y\in\mathbb N\) is read as the position of the rightmost red cell relative to the same origin \(x_0\). In this case output \(y=20\).
    \emph{Update:} the transition rule is a lookup table over radius-1 neighbourhoods, applied synchronously on all the tape. 
    \emph{Halting:} triggered by a relative count of cell values (Explained in \cref{sec:em43}).}
    \label{fig:ded}
\end{figure}

Both the latent state and the interface admit far more general forms than this example. The latent state may be unbounded in space, as in a Turing machine tape, \(s\in\{0,1\}^{\mathbb N}\), or in the grid of the Game of Life CA, \(s\in\{0,1\}^{\mathbb Z^2}\). Encoding and decoding are similarly general. Information can be injected either at fixed positions of the state space, called \emph{ports} (Figure~\ref{fig:em-loop}), or at input-dependent locations determined by some computable rule (Figure~\ref{fig:ded}).

Emergent Models formalize this picture through a generalized automaton: a discrete-time dynamical system with a halting condition. Equipping the automaton with an interface (encoder and decoder) turns the dynamical system into a model. The following subsections make this construction mathematically rigorous.

\subsection{Generalised Automaton}
The topology of the latent state space \(\mathcal{S}\) is abstracted as a graph to keep the formalism as general as possible.

Let \(G=(V,E)\), where \(V\) is the set of vertices (nodes) and \(E\) is the set of edges.

Each node of the graph carries a value in a set \(W\), typically binary \(\{0,1\}\), finite discrete \(\{0,1,2,\dots,n\}\), or continuous \(\mathbb R\). The state space \(\mathcal S\) is the set of all global configurations, that is, all assignments of a value in \(W\) to each node, \(\mathcal S := W^V = \{\, s \mid s:V\to W \,\}\). The edges define the neighborhood structure, but do not play a direct role in the formalism, as explained below.

\begin{figure}[H]
    \centering
    \includegraphics[width=1\linewidth]{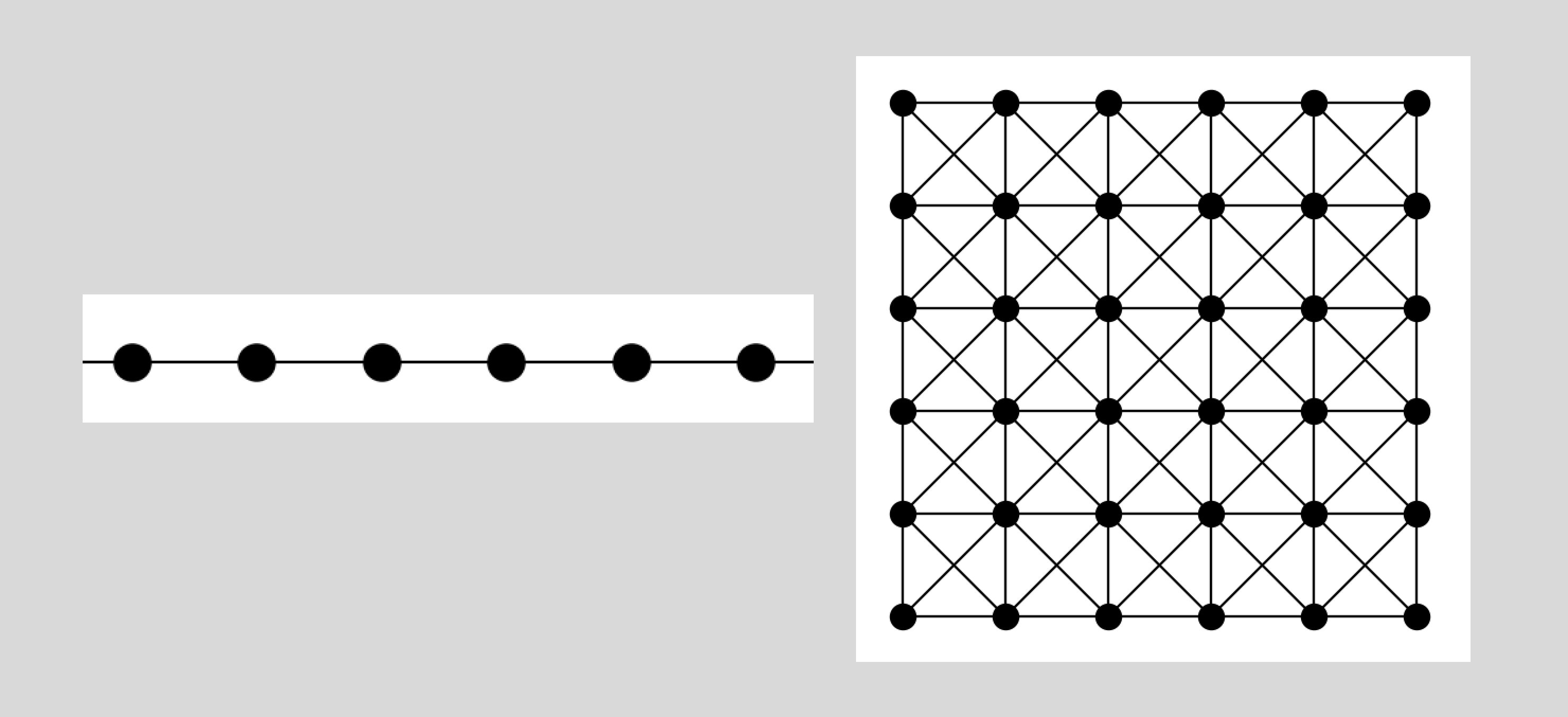}
    \caption{Graphs: 1-d unbounded tape (on the left) with radius-1 interaction ; 2-d finite lattice (on the right) with Moore neighborhood. Dots represent nodes, while lines connecting the nodes are the edges.}
    \label{fig:graphs}
\end{figure}

Typical choices are (i) a one-dimensional unbounded tape, as in Turing machines or 1-d CAs, with \(V=\mathbb N\) and finite alphabet \(W=\{0,1,\dots,n\}\), or (ii) a two-dimensional finite lattice, with \(V=\{0,1,\dots,n\}^2\) and cell values in \(\mathbb R\) or \(\mathbb Q\) (Figure~\ref{fig:graphs}); example (i) is instantiated in Figure~\ref{fig:ded}, and example (ii) in Figure~\ref{fig:em-loop}. These choices affect expressivity: to support unbounded memory capacity (a necessary condition for Turing completeness) the system must have either an unbounded topology, such as an infinite tape, or a finite tape or grid with an infinite value set \(W\), for example rationals or reals at unbounded precision.

A transition function \(f:\mathcal S\to\mathcal S\) acts on the state and performs the one-step update, which we call a \emph{microstep}, so that \(s_{t+1}=f(s_t)\). A halting operator \(H:\mathcal S\to\{0,1\}\) is checked at each microstep, and if \(H(s_t)=1\) the iteration stops. Starting from an initial configuration \(s_0\in\mathcal S\), the system evolves until halting. We define the halting time as \(T(s_0):=\min\{\,t\in\mathbb N \mid H(s_t)=1\,\}\), when this exists and is finite, and undefined otherwise. If \(T(s_0)\) exists, the terminal state is \(s_T:=s_{T(s_0)}\).

The transition function \(f\) is \emph{local} if there exists a function \(F\) such that for every \(s\in\mathcal S\) and every vertex \(v\in V\),\((f(s))(v) = F\bigl(s|_{N(v)}\bigr)\), where \(s|_{N(v)}\) denotes the restriction of \(s\) to the neighborhood \(N(v)\), typically induced by the edges.  In unbounded topologies (e.g. an infinite cellular-automaton lattice), we typically assume one value \(b \in W\) to be quiescent/blank, i.e. stable under the local rule: \(F(b,\dots,b)=b\), so that a blank cell remains blank whenever its neighborhood is composed only of blank cells.
Importantly, \(f\) is the global transition operator acting synchronously on the whole state space; locality means that this global update can be realized by a local rule \(F\) applied to each neighborhood. In this formulation, the edge structure need not be modeled separately: any dependence on adjacency is already absorbed into the definition of \(f\).

The transition function can be realized in many ways, such as lookup tables in discrete cellular automata, neural networks in neural cellular automata, polynomial maps, and other computable rules.

A generalized automaton is a tuple \(\mathcal A=(\mathcal S,f,H)\) that, starting from an initial condition \(s_0\), generates a trajectory \(s_0,s_1,s_2,\dots\) and, if halting occurs, terminates at a state \(s_T\). Many computational devices can be abstracted in this way, including Finite State Machines, Turing machines (see Appendix~\ref{app:tm-em}), CAs, NCAs, and some forms of recurrent neural networks.

\subsection{The model}
An Emergent Model is a tuple \(M= (\mathcal S,f,H,E,D,\oplus)\), and is built by equipping a generalized automaton with an input-output interface.

Let \(E:\mathcal X\to\mathcal S\) be an encoder function, mapping from an input space \(\mathcal X\) into the latent space \(\mathcal S\), and let \(D:\mathcal S\to\mathcal Y\) be a decoder, reading an output in \(\mathcal Y\) from the latent space \(\mathcal S\).

The automaton starts from a non-perturbed \textit{(np)} initial state \(s_{0,np} = p\in\mathcal S\), which we call the \emph{program}. 
Given an input \(x\in\mathcal X\), the encoder produces a state \(E(x)\in\mathcal S\). The actual (perturbed) initial state is then obtained by combining the program with the encoded input:
\[
s_0 = s_{0,pert} = p \oplus E(x)
\]
where \(\oplus:\mathcal S\times\mathcal S\to\mathcal S\) is a simple total computable combining operator, typically given by sum, overwrite at particular locations or similar operations.

\begin{figure}[H]
    \centering
    \includegraphics[width=1\linewidth]{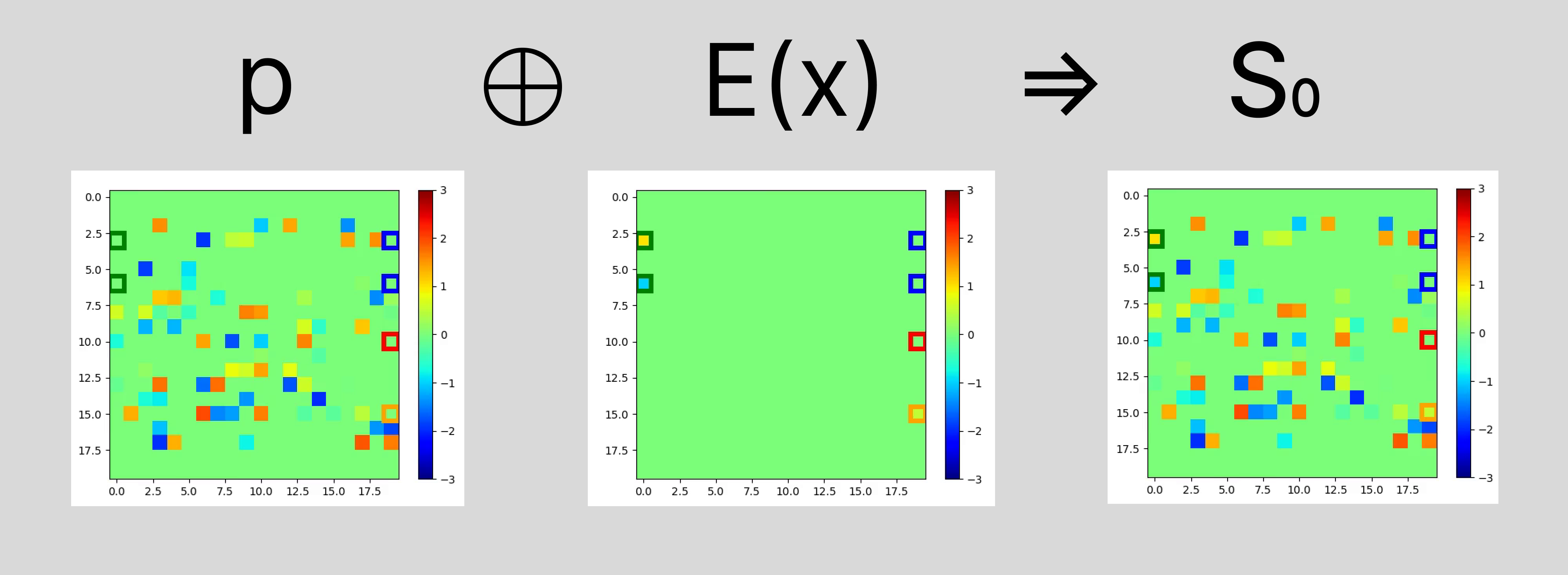}
    \caption{Construction of the initial state \(s_0 = p\oplus E(x)\) on the same
    two-dimensional continuous-valued lattice of \cref{fig:em-loop}; here \(\oplus\) is a sum.
    \emph{Left:} the program \(p\), non-perturbed initial state.
    \emph{Centre:} the encoded input \(E(x)\), nonzero only at the input ports and
    quiescent elsewhere. 
    \emph{Right:} their sum, giving the perturbed initial state
    \(s_0\) from which the dynamics are iterated.}
    \label{fig:ic-construction}
\end{figure}

From this point on, \emph{program} \(p\) always indicates the non-perturbed initial state, while \emph{initial state} refers to the perturbed initial state \(s_0=p\oplus E(x)\). An example construction of the initial state by combination is shown in \cref{fig:ic-construction}.

The program \(p\) should not be understood as a symbolic algorithm explicitly compiled into the substrate, as in a programming language. It is instead a raw substrate-level configuration: a pattern in the state space that conditions the system's evolution so as to induce a particular function. In this sense it still captures what we take to be the essence of a \emph{program} (a state held in memory that steers the computation) but without the symbolic, human-written form that word usually implies.

Inference proceeds as follows: initialize the state with the program, encode the input, run the dynamics \(s_{t+1}=f(s_t)\) until halting, and decode the output from the terminal state. If the computation never halts, the output is undefined.

This induces the (partial) input-output map:
\[
\Phi(x)
=
D\!\Bigl(f^{\,T(p,x)}\bigl(p\oplus E(x)\bigr)\Bigr)
\]
whenever the halting time \(T(p,x)\) is defined.

In machine learning, a model is treated as a \emph{parametrized function} \(\Phi_\theta:\mathcal X\rightharpoonup\mathcal Y\), where the parameters \(\theta\in\Theta\) collect the components allowed to vary during training. In an Emergent Model, \(\theta\) may include the program \(p\), parts of the transition function \(f\), of the halting condition \(H\), and of the interface \(E,D,\oplus\).

Modeling then means finding parameters \(\theta\) whose induced map \emph{computes} a target function \(g:\mathcal X\to\mathcal Y\). This notion is deliberately stronger than the local-approximation property usually invoked in deep learning, which guarantees a fit only on a compact domain: computing a function means realizing it on its full, possibly open domain, reproducing its behaviour everywhere rather than interpolating its values on a bounded region (exactly in the discrete case, or to arbitrary precision in the continuous case).

\begin{definition}[Model computes a function]
Let \(\Phi_\theta:\mathcal X \rightharpoonup \mathcal Y\) be a model parametrized by \(\theta\), inducing a family of partial functions \(\{\Phi_\theta\}_{\theta\in\Theta}\). Let \(g:\mathcal X \rightharpoonup \mathcal Y\) be a target partial computable function, total computable on the domain \(\tilde{\mathcal X}\subseteq\mathcal X\).

If \(g\) is discrete, we say that the model \emph{computes} \(g\) if for some \(\theta\in\Theta\):
\[
\Phi_\theta(x)=g(x)\quad \forall x\in\tilde{\mathcal X}
\]
and \(\Phi_\theta(x)\) is undefined for \(x\notin\tilde{\mathcal X}\).

If \(g\) is continuous-valued, the notion becomes arbitrary-precision global approximation: the model \emph{computes} \(g\) if, for every output tolerance \(\varepsilon>0\), there exists \(\theta_\varepsilon\in\Theta\) such that, for every \(x\in\tilde{\mathcal X}\), a sufficiently precise finite representation \(x_{\mathrm{fin}}\) of \(x\) satisfies:
\[
\|\Phi_{\theta_\varepsilon}(x_{\mathrm{fin}})-g(x)\|<\varepsilon
\]
The tolerance \(\varepsilon\) is fixed over the whole output domain \(\mathcal{Y}\), while the required input precision may depend on \(x\). A more explicit formulation is given in Appendix~\ref{app:cont-comp}, where we also show that, in the asymptotic limit of input precision, uniform global approximation is obtained.
\end{definition}

Which functions a model can compute depends on both the substrate and the parametrization chosen. Different substrates and parametrizations therefore compute different classes of functions, presenting different degrees of expressivity. The maximal degree in the computational setting is Turing completeness, or computational universality: a single model able to represent every partial computable function by varying \(\theta\).

We theorize a special regime, called \emph{latent universality} (Section~\ref{sec:lu}), in which the transition function, state space topology, and interface are all fixed and the program alone determines the computation the model performs (\(\theta=p\)). In this regime, taking \(\mathcal X\) and \(\mathcal Y\) to be a general computable data format (e.g.\ arbitrary length binary strings \(\{0,1\}^*\)) and a universal automaton \(\mathcal A\), a single fixed interface \(E,D,\oplus\) suffices for all computable functions on that domain (see \Cref{thm:lu}). In non-universal constructions, including the practical implementations we tested, the transition function or parts of the interface may instead be trained jointly with the program.

The encoder and decoder must be simple, total computable, and compatible with the dynamics of the chosen automaton \(\mathcal A=(\mathcal S,f,H)\). Not every interface is admissible: some encodings produce representations the dynamics cannot process, and some decodings require output structures the dynamics cannot produce.

\subsection{Latent Universality}
\label{sec:lu}
Assuming the input/output spaces are binary strings \(\mathcal X=\mathcal Y=\{0,1\}^*\), we consider a regime in which an Emergent Model \(\mathsf M\) is kept with fixed components \((\mathcal S,f,H,E,D,\oplus)\), while arbitrary computations are realized by varying only the program \(p\in\mathcal S\). We call this \emph{latent universality}: the latent state alone can select the realization of any possible algorithm performed by the model.

Latent universality is not meant as a new universality claim at the level of computability theory, but as a reformulation of it in natural terms for machine learning on dynamical systems. In Universal Turing Machines (UTMs), arbitrary computable behavior is realized by a fixed transition function together with a fixed scheme for encoding program and input into the initial state of the tape \citep{sipser2012introduction,arora2009computational,jerabek_mathlogic_notes}. In the EM setting, the same structure appears as a fixed substrate $(\mathcal S,f,H)$ and interface $(E,D, \oplus)$, with the program state defining the computation of a particular function (see Appendix~\ref{app:utm-em}). 

The analogy should not be interpreted as requiring a compiler from symbolic algorithms to substrate programs: the program is treated as a trainable substrate-level variable, and learning consists of searching over raw program states that induce the desired behavior when paired with inputs. Inputs, by contrast, may be compiled: the encoder \(E\) syntactically transforms them into a fixed representation the machine can read, with no information about the program \(p\) nor about the target function to be realized. This asymmetry is what motivates keeping the program \(p\) and the encoded input \(E(x)\) separate and combining them afterwards, rather than compiling both jointly through a single map \(E(p,x)\): since the encoder acts only on \(x\) and not on \(p\), any program can be searched while inputs are presented to the substrate through a common, fixed interface. This formulation departs slightly from the standard pairing convention, which does not generally require program and input to be separable; several known universal Turing machines nonetheless admit this disjoint form, initializing the tape with the simulated machine description and the encoded input in distinct regions or tapes, with no interdependence between the two: the encoder of \(x\) has no information about \(p\), and the combination consists on a simple operation (e.g.\ overwrite or concatenation) that applies no semantic interpretation of either \(x\) or \(p\) (see Appendix~\ref{app:utm-em}).

We work in a strong-universality setting. The substrate has an unbounded topology (countably infinite \(V\)) and a finite discrete alphabet \(W\) with a quiescent (blank) symbol \(b\in W\). Program and encoded input are \emph{finite-support} states: all non-quiescent values are confined to a finite but arbitrarily large region, surrounded by an infinite quiescent background. For every program \(p\) and input \(x\), the combined initial state \(p\oplus E(x)\) has finite support as well.

\begin{theorem}[Latent Universality]
\label{thm:lu}
Let \(\Sigma=\{0,1\}\) and the input/output spaces be binary strings \(\mathcal X = \mathcal Y = \Sigma^*\).
There exists an Emergent Model: \(\mathsf M=(\mathcal S,f,H,E,D,\oplus)\), with \(\mathcal S = W^V\) for some countably infinite \(V\) and finite-discrete \(W\), such that for every partial computable function \(g:\Sigma^*\rightharpoonup\Sigma^*\), there exists a finite-support program state \(p\in\mathcal S\) for which \(\Phi_p\) computes \(g\).
\end{theorem}
In Appendix~\ref{app:lu-proof} we provide a proof sketch of this result and extend it to Conway's Game of Life CA through Rendell's construction (\Cref{lem:gol-lu}).

Using \(\mathcal X,\mathcal Y=\{0,1\}^*\) is without loss of generality, since binary computable functions already capture the full class of computable functions. Importantly, other computable domains must be reduced to binary strings through an effective binary expansion, adding a small outer layer to the interface that converts that data format into binary.

\begin{figure}[H]
    \centering
    \includegraphics[width=0.7\linewidth]{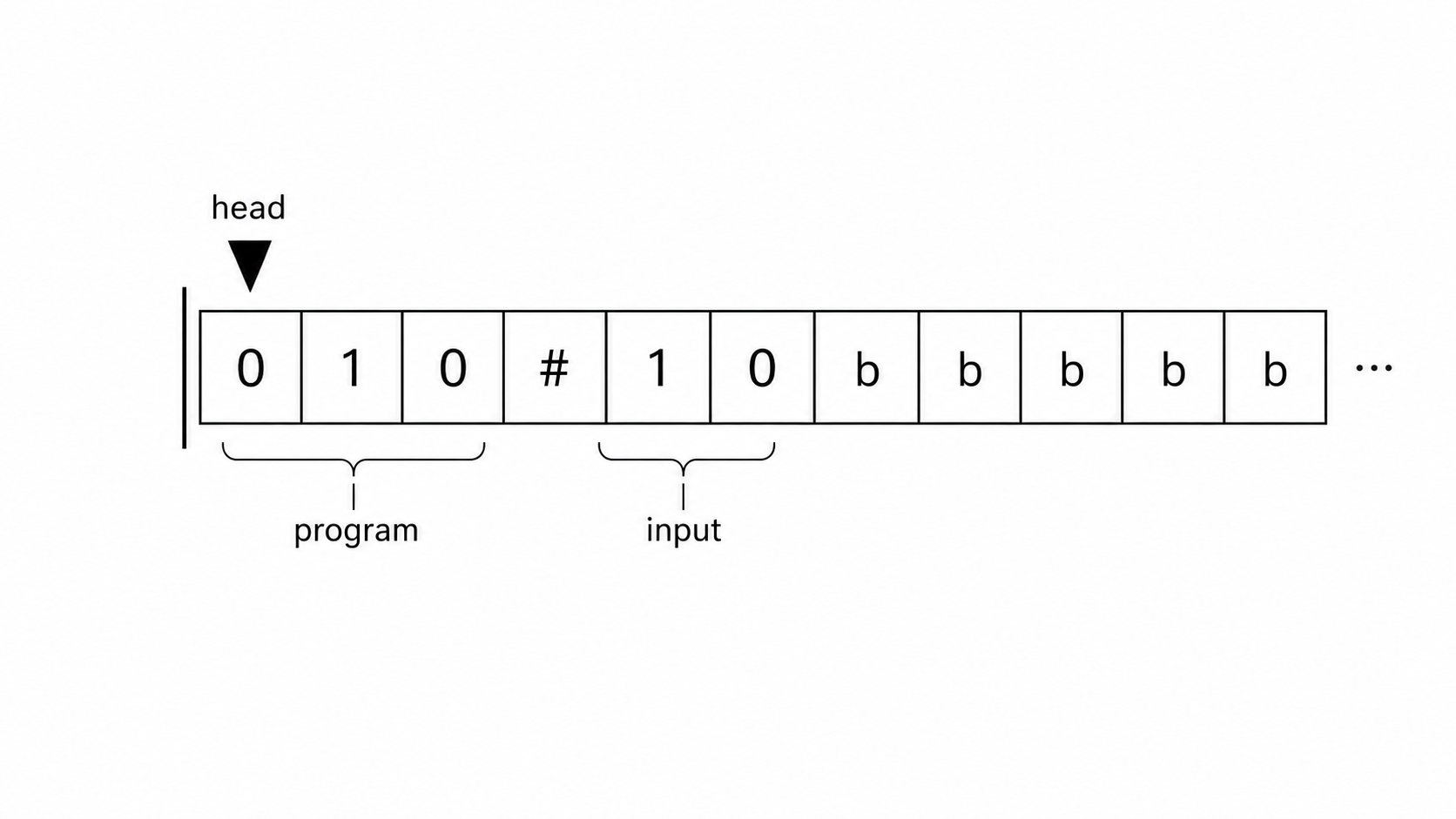}
    \caption{Didactic TM example: program string \(\rho\), separator \(\#\), and input string \(x\) on the tape, forming the initial state \(\rho\,\#\,x\, b^{\infty}\).}
    \label{fig:tm-didactical}
\end{figure}

On a one-sided, one-dimensional tape (\(V=\mathbb N\)), a finite-support state is exactly a finite string over \(W\) followed by an infinite quiescent background: the program takes the form \(p=\rho\, b^{\infty}\) and the encoded input \(E(x)=\chi\, b^{\infty}\), with \(\rho,\chi\in W^{*}\). As a didactic example, consider a Turing machine with alphabet \(\Gamma=\{b,0,1,\#\}\), where \(\{0,1\}\) are used for program, input, and output, \(b\) is the blank, and \(\#\) is a separator. The encoder writes the binary input \(x\in\{0,1\}^*\) verbatim, followed by an infinite quiescent background, \(E(x)=x\, b^{\infty}\). The combining operator strips the quiescent backgrounds, concatenates the two finite strings around the separator, and re-pads with an infinite quiescent background, giving \(p\oplus E(x)=\rho\,\#\,x\, b^{\infty}\) (\cref{fig:tm-didactical}).

The same structure extends beyond tapes. On a two-dimensional grid, the program and the encoded input are both full-grid states, each non-quiescent only on its own finite region, and \(\oplus\) merges them into a single state on the same board, as in the Game of Life construction of Lemma~\ref{lem:gol-lu}. Treating program and encoded input as states (\(p,E(x)\in\mathcal S\)), rather than strings (\(\rho,\chi\in W^{*}\)), is what makes this possible: the construction applies to any generalized automaton, over arbitrary graph topologies, and the classical string picture is recovered as the one-dimensional special case.

The flexibility of the paradigm leaves open further constructions: with a continuous value set \(W\), for instance, the same scheme may provide a substrate for analog computing. We currently offer no proof or formal guarantee in that direction.

\subsection{Sequential Operation, State Retention and Meta Learning}
\label{sec:retention}
So far, the EM formalism defines a predictor acting on a single input, \(y=\Phi_p(x)\). Many tasks, however, require sequential operation on an input stream \((x_i)_{i\ge 0}\) with memory retained across successive predictions. This introduces two time scales: microsteps \(t\), indexing the internal reasoning steps of the model within a single prediction, and macrosteps \(i\), indexing successive data points on the stream.

A naive rule to pass information to the next macrostep would be to set the next program equal to the previous terminal state, \(p_{i+1}=s_{T_i,i}\). In general this may fail, since \(s_{T_i,i}\) already satisfies the halting condition and would make the next prediction halt immediately. We therefore introduce a \emph{retention operator} \(R:\mathcal S\to\mathcal S\), whose role is to transform the terminal state into a valid program for the next macrostep:
\[
p_{i+1}=R(s_{T_i,i})
\]
The operator \(R\) should preserve the information that must persist across macrosteps, while restoring the control conditions needed for a new computation.

For example, in a Turing Machine, \(R\) could reset the head state from the halting state \(q_h\) to the initial state \(q_0\), reset head position to index 0, while preserving the tape content. Such a reset is valid only if the retained state does not prevent the machine from starting and performing the next computation correctly. Retention may require a substrate and interface designed so that persistent memory can coexist with repeated computation and not every latent-universal construction is necessarily compatible with this regime.

The program at first macrostep is initialized as \(p_0\in\mathcal S\), so at every macrostep \(i\) the model is initialized as \(s_{0,i}=p_i\oplus E(x_i)\), evolves by microsteps \(s_{t+1,i}=f(s_{t,i})\) until halting at time \(T_i\), outputs \(y_i=D(s_{T_i,i})\), and sets the next program to \(p_{i+1}=R(s_{T_i,i})\). If some prediction never halts, the sequential process fails. In general, this cannot be avoided without restricting the class of computations the model can express.

In control or reinforcement-learning settings, it is natural to augment each macrostep input with reward information, writing \(x'_i=[x_i,r_i]\), where \(r_i\) is the instantaneous reward from the previous macrostep.

Retained state is not merely passive memory. Because each new computation starts from the retained program \(p_{i+1}=R(s_{T_i,i})\), the stream of inputs can influence the future computation itself. In latent-universal settings, universality implies that the substrate can in principle represent any algorithm, while the latent nature of the program makes that algorithm part of the mutable state itself, suggesting the following conjecture: given sufficient memory and computation time, and providing the reward as an input, there exist some EMs that implement internally: (i) a prediction subroutine, which maps the current input to an output, and (ii) an update subroutine, which uses reward signal to modify a designated sub-portion of the program. In this view, the model could not only execute a prediction but also a feedback-driven self-improvement procedure acting on its own latent program, realizing a form of meta-learning \citep{vanschoren2018metalearningsurvey}. However, this remains speculative and requires further analysis, especially to determine the maximum degree of self-editing plasticity a substrate can support and the implications for robustness.

\subsection{Extensions and Relaxations}
The Emergent Model theoretical framework extends naturally to continuous-time dynamical systems, and in particular to continuous media governed by local field dynamics.

At the most general level, one may replace the discrete transition \(s_{t+1}=f(s_t)\) with a continuous-time evolution law:
\[
\frac{ds}{dt} = f(s)
\]
This ODE form is an abstract state-space description and does not, by itself, impose any notion of locality.
A more structured extension is obtained when the state is a field over an Euclidean spatial domain \(\Omega \subseteq \mathbb R^n\), so that:
\[
s(\cdot,t): \Omega \to \mathbb R^m
\]
where the \(m\) channels may represent multiple interacting quantities. The dynamics are then specified by a local evolution law of PDE type, for example:
\[
\partial_t s = F\!\bigl(s,\nabla s,\nabla^2 s,\dots\bigr)
\]
This yields a natural notion of continuous-space computation, encompassing substrates such as reaction-diffusion systems and other fluid-like media.

Under these extensions, the interface formalism remains unchanged: the encoder \(E\), decoder \(D\), halting \(H\), and combining operator \(\oplus\) are still defined as maps on the state space. The halting time is defined analogously as:
\[
T(s_0):=\inf\{\,t\ge 0 \mid H(s(t))=1\,\}
\]
whenever this quantity exists and is finite.

Another natural relaxation is allowing stochastic evolution rules, where the dynamics include random fluctuations or noise, so the same initial state may generate different trajectories. The induced output is then stochastic.

A more practical relaxation is to cap, or even fix the internal computation time. Instead of running the dynamics until an explicit halting condition is met, one may apply the transition rule for a fixed time budget \(T\). This reduces expressivity but also removes the possibility of non-halting predictions, makes inference easier to parallelize on hardware, and can simplify state retention.
Under suitable conditions, this allows identity state retention (\(R(s)=s\)) : the final state can be used directly as the program for the next macrostep, \(p_{i+1} \leftarrow s_{T,i}\), without requiring to cleanup a halting region.

A further, mostly notational relaxation is to reframe the state initialization in pseudocode. Instead of explicitly forming \(p\oplus E(x)\), one practically sets the initial state as the program and then overwrites the encoded input into a designated subspace:
\[
\begin{array}{l}
\texttt{s0 <- p} \\
\texttt{I <- InputSubspace(s0, x)} \\
\texttt{s0[I] <- E(x)}
\end{array}
\]

Here \(I\) denotes the input subspace, which may vary in size and location. The encoder becomes an operator acting on this subspace, and the assignment \(\texttt{s0[I] <- E(x)}\) represents an overwrite operation that injects the encoded input into that region, while the rest of \(s_0\) is left unchanged.

For example, if the state space is a vector and the input is \(x=[x_1,x_2]\), the encoder may simply write the input into fixed coordinates:
\[
\begin{array}{l}
\texttt{s0 <- p} \\
\texttt{s0[2:4] <- [x1, x2]}
\end{array}
\]

Under this procedural view, an Emergent Model can be represented in pseudocode as follows.

For non-sequential tasks, without retention, a single prediction is:
\[
\begin{array}{l}
\textbf{Input } x \\
\textbf{Program } p \\
\\[-0.6em]
\texttt{s <- p} \\
\texttt{I <- InputSubspace(s, x)} \\
\texttt{s[I] <- E(x)} \\
\textbf{for } \texttt{t in microsteps} \textbf{ do} \\
\qquad \texttt{s <- f(s)} \\
\texttt{y <- D(s)} \\
\end{array}
\]

On sequential tasks, with retention:

\[
\begin{array}{l}
\textbf{Input stream } (x_i)_{i\in\mathrm{macrosteps}} \\
\textbf{Program } p \\
\\[-0.6em]
\texttt{s <- p} \\
\textbf{for } \texttt{i in macrosteps} \textbf{ do} \\
\qquad \texttt{I <- InputSubspace(s, x\_i)} \\
\qquad \texttt{s[I] <- E(x\_i)} \\
\qquad \textbf{for } \texttt{t in microsteps} \textbf{ do} \\
\qquad\qquad \texttt{s <- f(s)} \\
\qquad \texttt{y\_i <- D(s)} \\
\qquad \texttt{s <- R(s)} \\
\end{array}
\]

\subsection{Training}
After choosing which components are trainable, optimizing an Emergent Model means searching for configurations that make the induced computation solve a desired task. In supervised settings, this amounts to minimizing a prediction error; in control and reinforcement-learning settings, to maximizing expected cumulative reward \citep{goodfellow2016deep,sutton2018reinforcement}.

Let \(\theta\) denote the trainable components and \(\hat y=\Phi_\theta(x)\) denote the model's output. In supervised learning we solve:
\[
\theta^\star \in \arg\min_\theta \mathbb E[\ell(\hat y,y)],
\]
while in control tasks we solve:
\[
\theta^\star \in \arg\max_\theta \mathbb E\!\left[\sum_{i=0}^{N-1} r_i\right].
\]

In the most general case, \(\theta\) may include the program \(p\), the transition rule \(f\), the halting condition \(H\), and parts of the interface \(E,D,\oplus\). 
It is useful to distinguish between two kinds of trainable components. We call \emph{soft parameters} the ones belonging to the program $p$. These are latent and dynamical, and may in principle change during inference under state retention. We call \emph{hard parameters} the trainable components belonging to the transition function, halting and interface. These play a hardware-like role: they can be optimized during training, but remain fixed during inference.

In latent-universal settings only the soft parameters are trained for a particular task, while the hard ones \((\mathcal S,f,H,E,D,\oplus)\) remain fixed and reusable across tasks.

Crucially, the search is guided only by the decoded output: the intermediate microstep trajectory $s_0,\dots,s_T$ is left unsupervised, so any internal algorithmic structure that emerges is a byproduct of optimizing on the output error/reward alone.

Although many training algorithms can be employed, EM dynamics are strongly recursive, nonconvex, and often discrete, making gradient-based methods impractical or impossible to use. Evolutionary algorithms are therefore a natural choice, especially in the most open-ended scenarios. In latent-universal settings, any computable function is realized by some finite-support initial state \(p \in \mathcal S\). In one-dimensional substrates, such as Turing machines, the program state can be represented as a finite program string \(\rho\), followed by an infinite quiescent background \(b\), i.e. \(p=\rho b^\infty\). Training is then viewed as a form of evolutionary program synthesis over variable-length strings, where mutation allows insertion and deletion of elements \citep{gulwani2017program,koza1992genetic}.

\begin{lemma}[Searchability]
\label{lem:program-enumeration}
Let $\mathsf M$ be latent universal and assume programs are finite-support states, i.e.\ $p=\rho\,b^\infty$ with $\rho\in\{0,1\}^*$.
Then for every partial computable function $g:\{0,1\}^*\rightharpoonup\{0,1\}^*$ there exists such a finite string $\rho_g$ that allows the computation of $g$.
In particular, since $\{0,1\}^*$ is countable, enumerating programs $\rho$ (or sampling them at
random with variable length) will eventually hit a program that computes $g$.
\end{lemma}

The searchability lemma clarifies why finite-support programs are essential for treating latent-universal EMs as machine-learning models. If the target program has finite support, then it can be searched by enumeration, random sampling or by a finite sequence of edits (insertions, deletions, mutations). In this sense, EM training is viewed as \emph{emergent program synthesis}: rather than designing an algorithm at a high level and compiling it into the substrate, one searches over substrate-level initial states until a configuration inducing the desired behavior is found. The lemma provides only an asymptotic existence condition: a suitable program exists and is reachable in principle, but the search may be arbitrarily long and no efficiency guarantee is implied.

\section{Experiments}
\label{sec:experiments}
The experiments described below are intended to validate the Emergent Model framework and demonstrate it is non-vacuous, rather than to benchmark it against optimized task-specific approaches. None of the tested substrate-interface combinations is shown to be latent-universal; we favor evaluating a broad zoo of minimal models over engineering one toward the full theory. The goal is to test whether simple physical-like substrates, equipped with minimal input-output interfaces and trained by evolutionary search, can exhibit useful modeling capabilities, extrapolation, control, and adaptation.

Across experiments, we use generational population-based genetic algorithms (GAs) to optimize each model's trainable components, such as program state, transition-rule parameters, or interface parameters. Each generation evaluates candidate models, selects high-performing individuals through tournament selection, and forms the next population through elitism, sparse crossover, and mutation. This provides a simple black-box training method for discrete, non-differentiable, and strongly recurrent substrates. The exact GA variant and fitness objective vary slightly across tasks, and we report representative training and evaluation runs rather than full multi-seed statistical analyses.
In some control experiments, we apply state retention across macrosteps, episodes, and generations, so that inference and environment interactions modify the latent state that will be reused in subsequent evaluations. Thus the soft parameters are updated by the model's own dynamics and interaction history. When a parent is selected, crossover and mutation are applied only to its hard parameters, while its final latent state is copied directly to the offspring. We refer to this convention as \emph{Lamarckian-style state inheritance}: advantageous state changes can be inherited, even though the state is mutated by inference dynamics rather than by external noise.

The experiments are organized by task family: Arithmetic tasks test exact rule learning and extrapolation in discrete domains; CartPole tests simple closed-loop control; and Meta-Life tests a more complex control objective, with online policy adaptation.

\subsection{Arithmetic tasks}
\label{sec:exp-arithmetic}
We first evaluate Emergent Models on simple integer arithmetic tasks. Arithmetics is a natural test case for this framework because it requires learning an exact symbolic rule and extrapolating beyond the finite training range. Discrete local cellular automata are especially well suited to this setting, where inputs and outputs are integers, and computation can be represented through spatial interactions over a discrete latent state.

The learned program is fixed across datapoints of a task. For each input, the automaton is reinitialized from the program, then the input is encoded into the state, and the system is run until halting. No terminal state is retained, since these are non-sequential regression tasks.

\subsection{EM43}
\label{sec:em43}
We ran minimal experiments to test whether EM43, an Emergent Model made from a minimal discrete cellular automata, can learn simple arithmetic functions and extrapolate far beyond the training range. Training is performed jointly on the program and the rule for each of the following tasks: \(x+1\), \(2x\), \(\mathrm{round}(x/3)\), \(x \bmod 4\), \(x\cdot(1 + x\bmod 2)\) and \(a\cdot b\).  These tasks are respectively simple linear/affine expressions, periodic expressions with nontrivial asymptotic structure where extrapolation is typically difficult in machine learning, and finally a two-integer product task.

EM43 is a one-dimensional cellular automaton with one-sided tape, radius 1 and 4 values per cell, a position-based input-output interface and a majority-based halting rule. The value alphabet is \(W=\{0,1,2,3\}\), where \(0\) (white) is the quiescent background, \(1\) (black) a generic active cell, \(2\) (red) the input-output marker, and \(3\) (blue) the control value used for halting. A local lookup-table (LUT) rule \(F\) maps each radius-1 neighborhood \((s_t[j-1],s_t[j],s_t[j+1])\) to the updated value of the central cell \(s_{t+1}[j]\); the global update function \(f\) is obtained by applying \(F\) synchronously across the tape. To introduce minimal structure and reduce the search space, we fix some LUT entries: \(000 \mapsto 0\) to preserve quiescence, and \(020 \mapsto 2\), \(200 \mapsto 0\), \(002 \mapsto 0\) to preserve stability of the input marker when surrounded by quiescent cells.

Given a global automaton state \(s_t\), let \(\#\mathrm{alive}\) denote the count of cells with non-quiescent values \((s_t[j]\neq 0)\), and \(\#\mathrm{control}\) the count of cells assuming the control value \((s_t[j]=3)\).
The computation halts when control cells are the majority among alive (non-quiescent) cells:
\[
H(s_t)=\mathbf{1}\!\left(\frac{\#\mathrm{control}}{\#\mathrm{alive}}\ge\frac{1}{2}\right)
\]
The tape at initial state \(s_0\) is divided into a trainable program region of length \(L_p\), a blank separator of length \(L_s\), and an encoding region of length \(L_e\), with total length \(L=L_p+L_s+L_e\). Let \(x_0=L_p+L_s\) be the origin at the start of the encoding region. An integer input \(x\in\mathbb N\) is encoded by placing a red cell at position \(x_0+x\). At halting, the output is decoded as the position of the rightmost red cell relative to the origin, \(y=x_{\mathrm{rr}}-x_0\). Figure~\ref{fig:round} shows a representative space-time plot of the automaton, illustrating the program and encoded input at \(t=0\), as well as the decoded output at halting time \(t=T\).
For the two-input task \(a \cdot b\), the pair \((a,b)\in\mathbb{N}^2\) is encoded by placing two markers at \(x_0+a\) and \(x_0+b\); since the task is commutative, their order is irrelevant.
During training, predictions are treated as invalid if they fail to halt within a computation time budget \(T_{\max}\), produce no decodable output, or hit the right tape boundary. During inference, we let the tape self-extend in length whenever active cells approach the right boundary, and impose no computation-time limit, making the device unbounded in space and time.

The present setting was adopted as a largely arbitrary, minimally engineered design. It should therefore be regarded as one among many possible automaton-interface combinations, and more effective ones may exist.

Training is performed on inputs in \([0,30]\), while extrapolation is evaluated on \([30,1000]\). For the two-integer multiplication task, training and evaluation follow the analogous bounded and extrapolation regimes over pairs \((a,b)\in\mathbb N^2\).

The learning algorithm is a generational island-based genetic algorithm with tournament selection, pointwise mutation, crossover, elitism, and island migration (hyperparameter ranges are reported in footnote\footnote{\begingroup\scriptsize\raggedright
GA hyperparameters varied slightly by task. Typical ranges were:
program mutation rate \(0.02\)--\(0.06\);
rule mutation rate \(0.01\)--\(0.04\);
program crossover rate \(0.03\)--\(0.07\);
rule crossover rate \(0.02\)--\(0.06\);
density penalty \(k_d=0.003\);
invalid-run penalty \(5\);
tournament size \(2\);
elite fraction \(0.05\);
random-immigrant fraction \(0.05\);
cross-island immigrant fraction \(0.01\);
cross-island migration every \(10\)--\(30\) generations.
\par\endgroup}). Tournament provides selective pressure, while the island structure promotes parallel exploration and occasional exchange between subpopulations, helping to maintain diversity and reduce premature convergence \citep{whitley1994tutorial}.

Fitness $F$ is derived from the dataset mean absolute error (MAE), together with a small density penalty \(k_d\) that promotes sparser programs and a penalty \(k_{invalid}\) for invalid computations: \(F=-\mathrm{MAE}-k_d d-k_{\mathrm{invalid}}p_{\mathrm{invalid}}\) , where \(d\) is the fraction of alive cells in the program region, \(p_{\mathrm{invalid}}\) is the fraction of invalid predictions in the dataset evaluated.
Note that fitness depends only on the final decoded output (i.e.\ the position of the rightmost marker at halting); the intermediate space-time evolution is never supervised. Any structured internal computation therefore emerges solely from minimizing the scalar error on the output.

For each task, independent training runs are performed using a fixed population of 20 islands with 2000 individuals each (40000 total), and a number of generations increasing with task complexity. The parametrization consists of a trainable program of length \(L_p\) together with a lookup table of 60 entries. As both program cells and rule entries take values in a 4-state alphabet, each parameter corresponds to 2 bits, allowing an estimate of the total information content of the learned model.

\subsubsection{Results}
After training, exact-match accuracy is evaluated on the training and test sets and averaged over runs. Table~\ref{tab:results_summary} summarizes the results.

\begin{table}[H]
\centering
\footnotesize
\setlength{\tabcolsep}{4pt}
\begin{tabular}{@{}lccccccc@{}}
\toprule
task & runs & gens & \(L_p\) & params & bits & \makecell{train\\acc} & \makecell{test\\acc} \\
\midrule
\(x+1\)                  & 8 & 20  & 10 & 70  & 140 & 100\%     & 100\%     \\
\(2x\)                   & 8 & 90  & 10 & 70  & 140 & 100\%     & 100\%     \\
\(x \bmod 4\)            & 8 & 150 & 10 & 70  & 140 & 100\%     & 100\%     \\
\(\mathrm{round}(x/3)\)  & 8 & 400 & 10 & 70  & 140 & 100\%     & 100\%     \\
\(x\cdot(1 + x\bmod 2)\) & 2 & 500 & 10 & 70  & 140 & 100\%     & 100\%     \\
\(a \cdot b\)            & 2 & 800 & 50 & 110 & 220 & $<$20\%    & $<$10\%   \\
\bottomrule
\end{tabular}
\caption{Performance of EM43 across different tasks, showing number of runs, generations, parameter size, and accuracy.}
\label{tab:results_summary}
\end{table}

Perfect extrapolation is observed, in all runs, for all one-input tasks considered.
This behavior is especially notable for periodic functions such as \(x \bmod 4\), having a nontrivial asymptotic structure. As discussed in the introduction, such behavior is not globally representable by FFNNs without task-specific engineering. The same holds for \(x\cdot(1 + x\bmod 2)\), which is an oscillatory function with linearly growing amplitude. By contrast, the two-input multiplication task \(a \cdot b\) shows poor performance on both training and test sets and does not converge within the available compute budget.

\begin{figure}[H]
    \centering
    \includegraphics[width=0.9\linewidth]{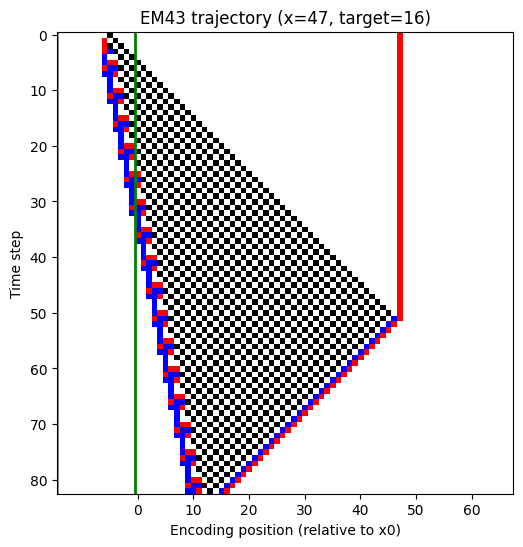}
    \caption{Space-time plot of EM43 computing \(\mathrm{round}(x/3)\) for \(x=47\), using the learned initial condition and update rule. The green line represents the origin of the encoding (\(x_0\)). The input is encoded as a red cell at position \((x_0 + x)\) in the initial condition, and the output is marked by the position of the rightmost red cell at halting time with respect to the origin.}
    \label{fig:round}
\end{figure}

\begin{figure}[H]
    \centering
    \includegraphics[width=1\linewidth]{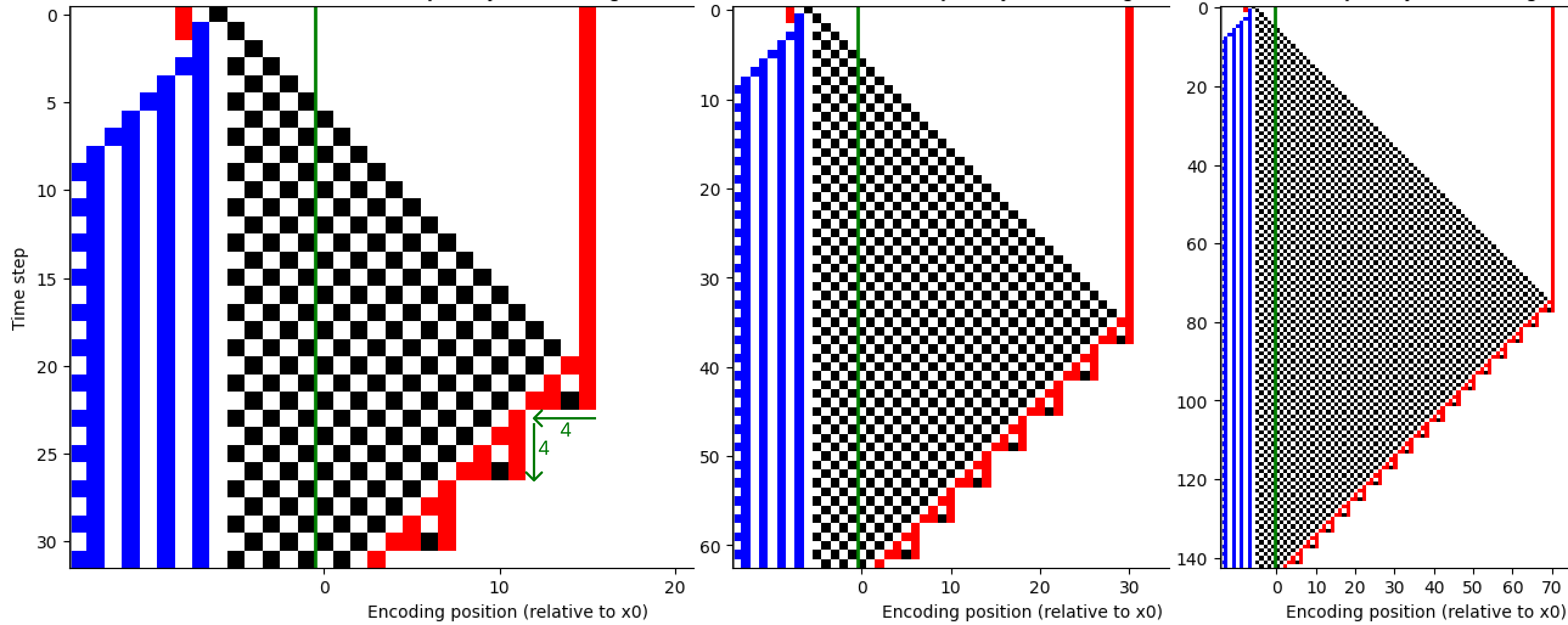}
    \caption{Spacetime plot of a model representing \(x\bmod4\), visualized for different inputs (15, 30 and 70).}
    \label{fig:mod}
\end{figure}

Plotting the space-time diagrams of configurations that generalize reveals an interesting phenomenon that we call \emph{geometric grokking}: exact computation is typically carried by regular and interpretable geometric patterns that remain stable across input scales. In the case of \(x \bmod 4\) (Figure~\ref{fig:mod}), the learned dynamics implement a simple mechanism that repeatedly shifts the red marker left by 4 and down by 4 until the system reaches a configuration in which blue cells are the majority and halting occurs. At that point, the red marker is found in the correct position to encode the output. This can be interpreted as an unconventional computing procedure realized through interactions in the latent space, and together with test accuracy results, seems suggesting the model has internalized the generative rule behind the data.

\begin{figure}[H]
    \centering
    \includegraphics[width=1\linewidth]{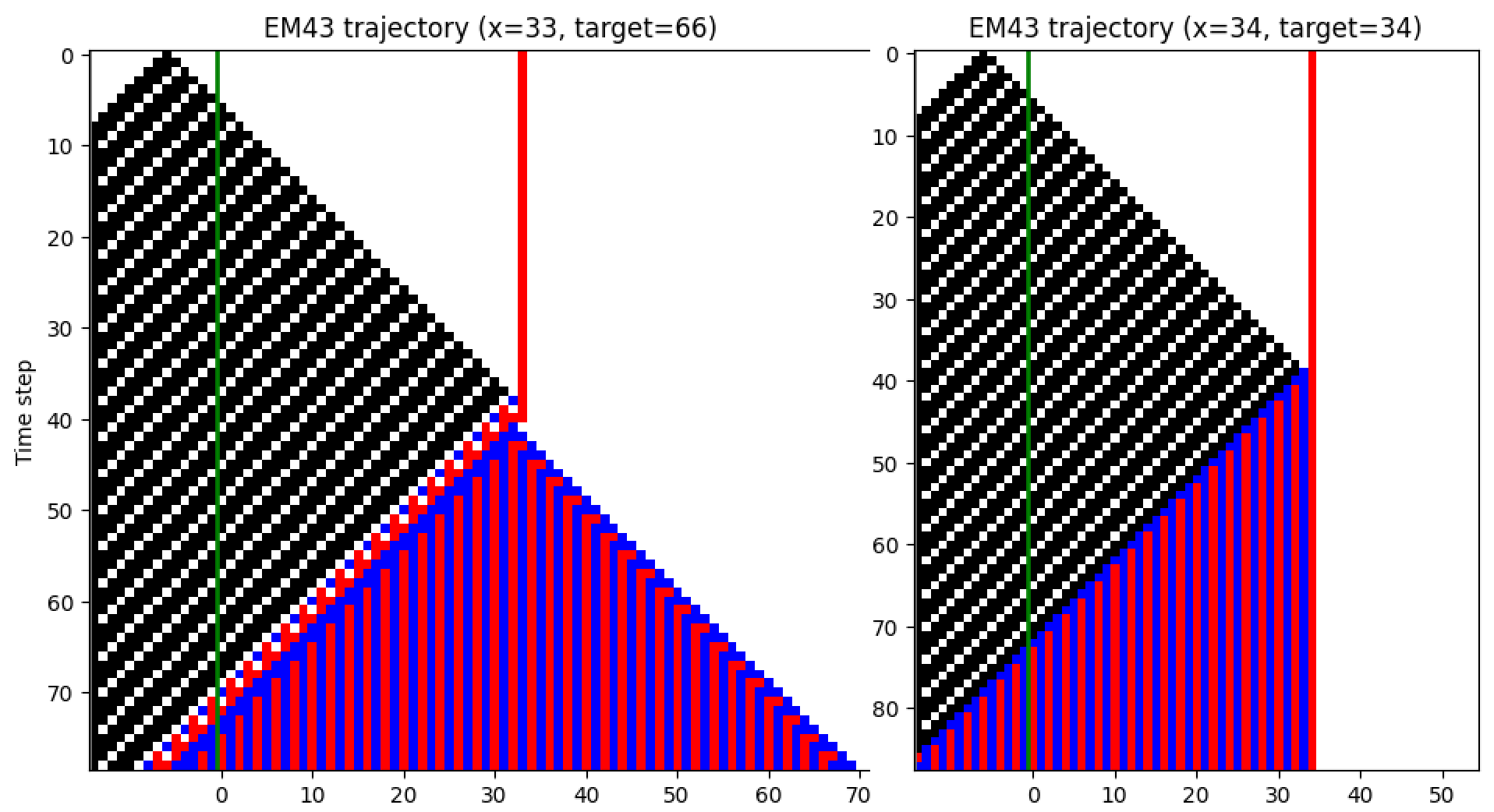}
    \caption{EM43 spacetime plot for \(x\cdot(1 + x\bmod 2)\). On the left for an odd input (33), on the right for an even input (34).}
    \label{fig:prod}
\end{figure}

The task \(x\cdot(1 + x\bmod 2)\) (see \cref{fig:prod}) is a product between a linearly growing function \(x\) and a periodic function \(1 + x\bmod 2\), resulting in the condition: if \(x\) is odd, return \(2x\); if \(x\) is even, return \(x\).

Different runs and tasks implement different geometric strategies, but they typically follow the same broad scheme: patterns propagate from the program region, collide with the input marker, and eventually trigger halting through changes in the active-cell population (see \Cref{fig:round} and \Cref{fig:prod}). 
One possible explanation of our results is that translation and time invariance of the update rule induce a homogeneous medium naturally suited to input scale generalization.

More complex tasks such as \(a \cdot b\) showed very slow convergence, and within the available training budget it was not possible to determine whether zero loss would eventually be reached.
At present, the source of failure cannot be clearly isolated. It may be due to insufficient training compute, hyperparameter choices (such as computation time and tape-length limits during training), interface limitations (the chosen encoding and decoding scheme), or insufficient expressivity of the automaton. Although the EM43 rule space likely contains weakly universal rules\footnote{\begingroup\scriptsize\raggedright
Given that weak universality already appears in the 2-state, radius-1 case through Rule 110.
\par\endgroup}, it remains unknown whether it contains a strongly universal rule and, if so, what interface would be compatible with it. Even under that optimistic assumption, search likely remains the dominant bottleneck, since genetic algorithms scale poorly in large combinatorial spaces and the number of possible programs grows as \(4^{L_p}\).

A notable aspect of this setting is the small size: 70 parameters carrying 2 bits each, corresponding to 140 bits of information. The setup is also data-efficient, being trained on only 30 integer examples while extrapolating to much larger ranges. This combination of parameter and data efficiency is broadly consistent with observations in the Neural Cellular Automata literature \citep{guichard2025arcnca,xu2025ncaarcagi}, and may suggest that local recursive computation is an important ingredient in the design of compact learning systems.

\subsection{GoL-EM}
\label{sec:exp-gol}

We next test the EM framework in Conway's Game of Life (GoL) on simple arithmetic tasks. The GoL rule is fixed, and training acts only on initial conditions. This matches the latent-program view from Section~\ref{sec:lu}: task-specific computation is encoded in the initial condition, not in the transition rule.

However, GoL's Turing completeness does not automatically transfer to this setup. Indeed the chosen input encoding, output decoding, and halting protocol can restrict access to universal constructions. Since our interface and halting policy is deliberately minimal and not proven to preserve universality, this experiment should be read as a small-scale test of initial-state optimization in a fixed cellular-automaton substrate, not as a demonstration of latent universality.

\subsubsection{Methodology}
We use a square GoL board as a fixed computational substrate. The automaton is binary-valued: \(1\) denotes a live cell and \(0\) denotes a dead cell, so the value set is \(W=\{0,1\}\). Ideally, the substrate is a square lattice extending indefinitely to the right and downward, while bounded on the left and upper sides; we write its vertex set as \(V=\mathbb N^2\), and the full state space as \(S=W^V=\{0,1\}^{\mathbb N^2}\). In practice, memory is finite, so it uses a finite square lattice of side length \(L\), nominally \(V=\{0,..,L-1\}^2\).

\begin{figure}[H]
    \centering
    \includegraphics[width=1\linewidth]{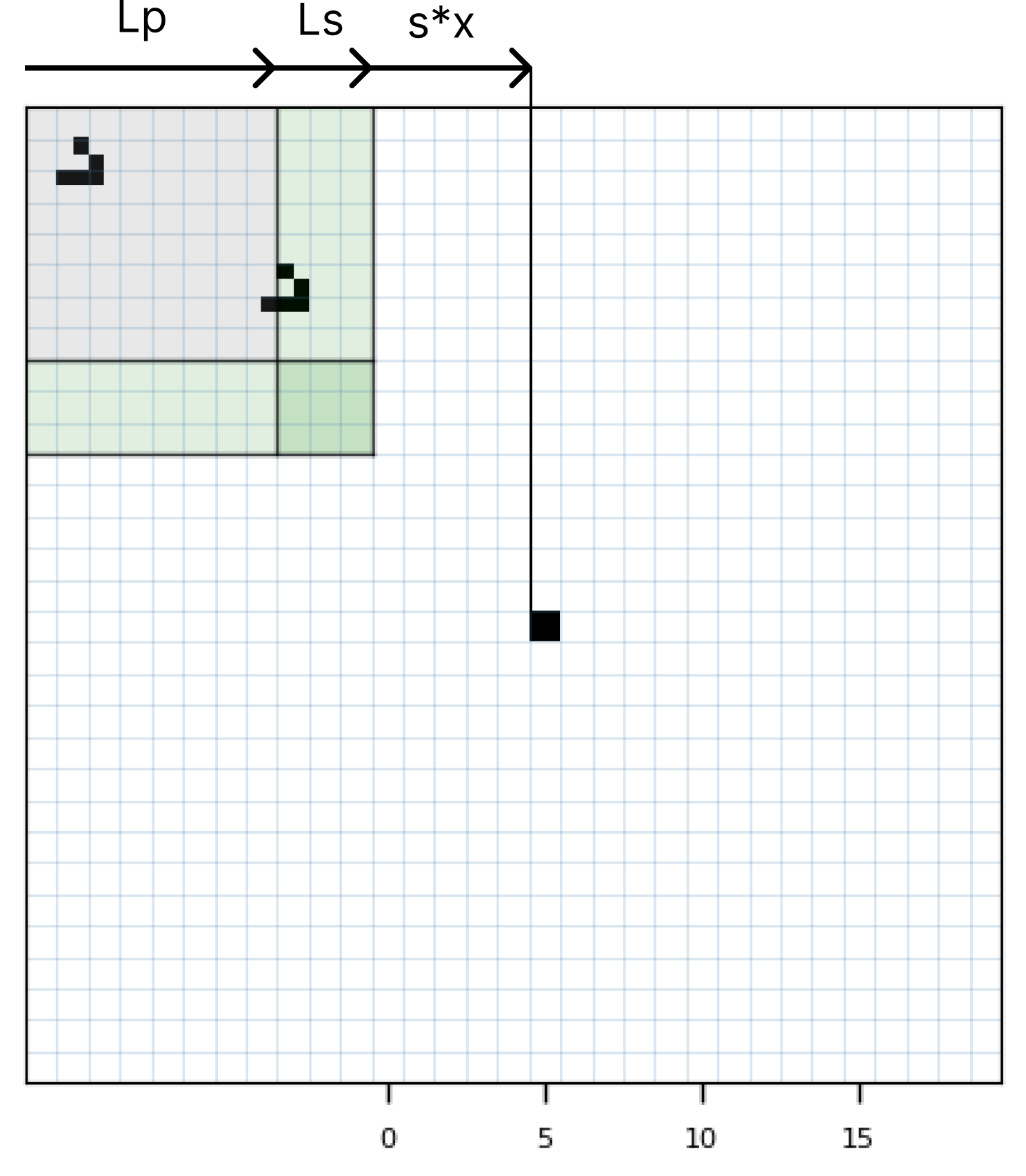}
    \caption{Automaton grid, here encoding the input $x=5$ and with a two-glider program. Each encoded integer corresponds to a \(2\times2\) block of automaton cells at a specific position; to represent this faithfully, the grid shown here is scaled by a factor of two relative to the underlying automaton lattice. Tick marks on the horizontal axis indicate the integer values represented by block positions, with respect to the origin.}
    \label{fig:gol-arith-encoding}
\end{figure}

The board is partitioned along each axis into three consecutive regions: the program region of side \(L_p\), a separator of width \(L_s\), and the encoding region. The separator keeps the program region disjoint from the area where inputs are written. Because the board is square and the layout is symmetric, the encoding region begins at the same offset on both axes, defining the encoding origin \((i_0,j_0)\) with \(i_0=j_0=L_p+L_s\). All inputs are encoded as positions relative to this origin. On a finite board of side length \(L\), the encoding region then spans \(L_e=L-j_0\) along each axis (See \cref{fig:gol-arith-encoding}).

To reduce the search burden, we do not optimize over arbitrary live/dead patterns in the program region; instead, we search over glider placements, using the upper-left corner of each glider as a trainable position variable. These gliders travel towards the input region and eventually interact with the encoded inputs.

Inputs are encoded positionally, on the diagonal of the encoding region. For an input value \(x\), we place a single \(2\times2\) live block (denoted \(\mathbf{1}_{2\times2}\)) at a position whose offset from the origin is proportional to \(x\), with stride \(s=2\). The block is placed along the diagonal: this layout is chosen because gliders travel diagonally in GoL, so a diagonal arrangement lets signals emitted from the program region reach the encoded input more easily. We use a \(2\times2\) block because it is a still life under the GoL rule, i.e. stable when isolated; the input therefore persists in place until those signals reach it.

\[
E(x):\space
s_{[(i_0+2x):(i_0+2x+2)\space,\;(j_0+2x):(j_0+2x+2)]}
\leftarrow
\mathbf 1_{2\times2}
\]

The state then evolves under the standard Conway's GoL update rule. Computation halts when the system reaches equilibrium/fixed point:

\[
H(s_t)=1
\quad\Longleftrightarrow\quad
f(s_t)=s_t
\]

At halting, the output is decoded by scanning the encoding region for isolated \(2\times2\) live blocks. Among all valid candidates, the leftmost block is selected, and its horizontal position relative to the encoding origin is converted back into a scalar using the same stride as in the input encoding. Unlike input, output blocks are not required to lie on the diagonal: only their horizontal position is used. Moreover, other live-cell debris that may remain elsewhere on the board is ignored by the decoder. This gives the dynamics more freedom to form a valid output and reduces the search burden.

Let \(j_{\mathrm{LB}}\) denote the horizontal position of the leftmost \(2\times2\) block (precisely, the column index of its upper-left corner), and let \(j_0\) be the encoding-origin column. The decoder returns:
\[
D(s_T)
=
\frac{j_{\mathrm{LB}}-j_0}{2}
\]

During training, if the halting condition is not reached within a maximum time budget \(T_{\max}\), the output is not decoded and a penalty is applied to the fitness. During evaluation, the same equilibrium-based halting condition is used, but the board is allowed to spatially self-extend to the right and downward whenever activity approaches the boundaries. Thus inference is not artificially limited by the initial board size. Evaluation is also allowed to run for an arbitrarily long time until equilibrium is reached.

We train the GoL program by minimizing prediction error while penalizing invalid computations:
\[
F
=
-\mathrm{MAE}
-
\lambda_{\mathrm{invalid}}\,p_{\mathrm{invalid}}
\]
Here \(\mathrm{MAE}\) is the mean absolute error on the training dataset, and \(p_{\mathrm{invalid}}\) is the fraction of examples that failed to produce a valid prediction. This includes both non-halting runs and halted runs with no decodable output block.

We optimize the program with two black-box search methods over glider positions in the program region: a genetic algorithm with mutation on glider positions, and plain random search that samples positions uniformly. The two reach almost identical results, so we report random search throughout, being the simpler. This near-equivalence is itself informative about the search landscape, and we return to it below.

\subsubsection{Results}
\begin{figure}[H]
    \centering
    \includegraphics[width=1\linewidth]{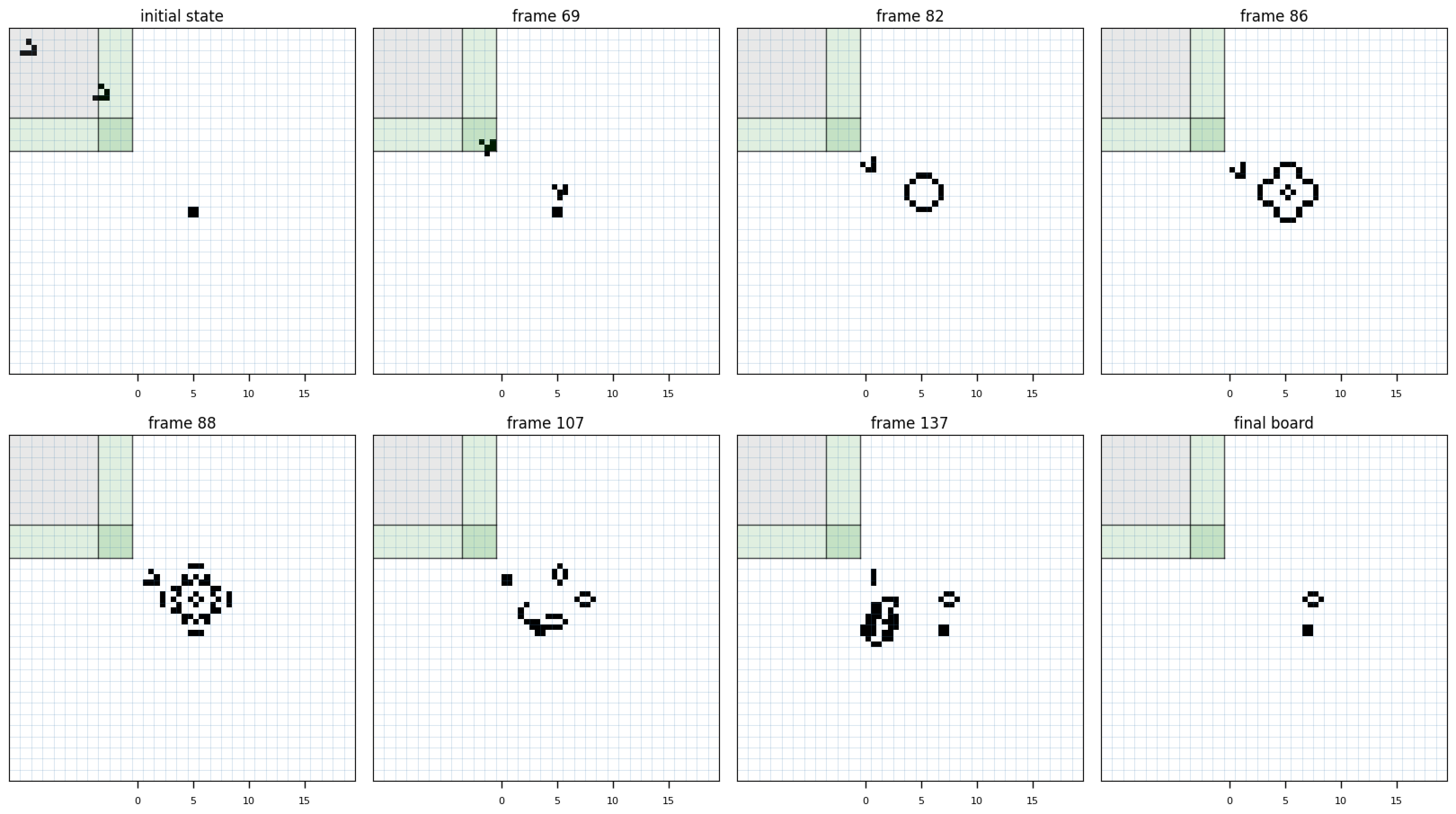}
    \caption{Observing a few steps of the trace computing $y=x+2$}
    \label{fig:gol-arith-trace}
\end{figure}

The GoL model successfully learned and generalized tasks of the family \(x\mapsto x+k\), such as \(x-2\), \(x-1\), \(x+1\), \(x+2\), and \(x+5\). By contrast, a more structured conditional task, returning \(x+1\) for even inputs and \(x-1\) for odd inputs, was not successfully learned.
The input ranges are given in \cref{tab:gol_train_eval_ranges}, the results in \cref{tab:gol_results}, and a representative model prediction in \cref{fig:gol-arith-trace}.

\begin{table}[H]
\centering
\footnotesize
\setlength{\tabcolsep}{4pt}
\begin{tabular}{@{}lcc@{}}
\toprule
task & train dataset \(x\)& test dataset \(x\)\\
\midrule
\(x-2\) & \([2,12]\) & \([12,200]\) \\
\(x-1\) & \([1,11]\) & \([11,200]\) \\
\(x+1\) & \([0,10]\) & \([10,200]\) \\
\(x+2\) & \([0,10]\) & \([10,200]\)\\
\(x+5\) & \([0,10]\) & \([10,200]\)\\
\makecell{\(x+1\), \(x\) even\\ \(x-1\), \(x\) odd} & \([0,10]\) & \([10,200]\)\\
\bottomrule
\end{tabular}
\caption{Training and evaluation input ranges for GoL arithmetic tasks. Lower bounds are chosen so that the target output is non-negative, since the positional decoder is not designed to represent negative integers.}
\label{tab:gol_train_eval_ranges}
\end{table}

\begin{table}[H]
\centering
\footnotesize
\setlength{\tabcolsep}{4pt}
\begin{tabular}{@{}lccccc@{}}
\toprule
task & \makecell{\(n_{\mathrm{evals}}\)} & \makecell{\(n_{\mathrm{gliders}}\)} & \(L_p\) & \makecell{train\\acc} & \makecell{test\\acc} \\
\midrule
\(x-2\) & 40k & 2 & 18 & 100\% & 100\% \\
\(x-1\) & 40k & 2 & 18 & 100\% & 100\% \\
\(x+1\) & 40k & 2 & 18 & 100\% & 100\% \\
\(x+2\) & 40k & 2 & 18 & 100\% & 100\% \\
\(x+5\) & 40k & 2 & 18 & 100\% & 100\% \\
\makecell{\(x+1\), \(x\) even\\ \(x-1\), \(x\) odd}
& 1M & 3 & 26 & 60\% & 50\% \\
\bottomrule
\end{tabular}
\caption{Performance of the GoL-based Emergent Model across tested tasks, trained by random search over glider positions. \(n_{\mathrm{evals}}\) is the search budget (number of candidate programs sampled and evaluated). Accuracies are exact-match rates.}
\label{tab:gol_results}
\end{table}

Several limitations remain. First, although GoL itself is universal under proper constructions (such as Rendell's \citep{rendell2016turing}), it is unknown whether the interface and halting condition chosen here preserve universality. Second, we observed that random search performs almost as efficiently as the genetic algorithm, suggesting that the search landscape is weakly structured, as the program space appears to lack a natural ordering or smoothness: small variations in glider positions do not seem to induce correspondingly small variations in fitness.

\subsection{Control tasks}
\label{sec:exp-control}

We next evaluate some Emergent Models in control settings. Differently from arithmetic tasks, the goal here is not extrapolation over unbounded ranges: observations and actions are typically bounded by the environment. Instead, it tests whether a simple substrate can support closed-loop behavior and robustness.

State retention becomes more relevant in this setting, because the controller acts on a stream of observations rather than on isolated datapoints. Instead, adaptive computation time through explicit halting is less central here than in static arithmetic tasks: in a control loop, computation is already distributed across macrosteps, since each new observation triggers another round of internal updates. For this reason, some of the controllers below use a fixed computation time budget \(T\) rather than a dynamic halting condition.

We consider two environments. The first is CartPole, a standard continuous-state control task in which the agent must balance a pole by applying forces to a cart. The second is Meta-Life, a simple 2-d spatial navigation and foraging environment, which we study in two variants: \emph{Meta-Life-Food}, where agents learn to move and collect food, and \emph{Meta-Life-Adapt}, where resources alternate between food and poison in phases of random duration: during a food phase, collecting any resource gives a positive reward; the world then switches to a poison phase, in which collecting a resource gives negative reward; it later switches back to food, and so on. The agent receives no a priori information about the current phase, and hence about whether resources are currently food or poison, since they are visually identical; it must therefore taste a resource, observe the resulting reward, infer the current phase from it, and adapt online, chasing resources during food phases and avoiding them during poison phases. Both variants share the same world and sensors, differing only in the reward assignation.

\subsection{CartPole}
\label{sec:exp-cartpole}

We use CartPole as a minimal closed-loop control task. At each environment macrostep \(i\), the controller observes the cart-pole state:
\[
x_i=(q_i,\dot q_i,\theta_i,\dot\theta_i)
\]
where \(q_i\) is cart position, \(\dot q_i\) cart velocity, \(\theta_i\) pole angle, and \(\dot\theta_i\) its angular velocity. The controller outputs a force action \(F_i\) applied to the cart.

We test two action formulations: a continuous force \(F_i\in[-F_{\max},F_{\max}]\), and a binary force \(F_i\in\{-F_{\max},+F_{\max}\}\).

A rollout is one CartPole episode, lasting at most \(N_{max}\) macrosteps. The objective is to keep the pole balanced for as long as possible within this time window, without exiting the allowed position range. At each macrostep, the agent receives reward \(+1\) if the cart remains within the allowed position range and the pole within the allowed angle range. If either condition is violated, the agent dies receiving no further reward, and the rollout is terminated. The return of rollout \(r\) is therefore:
\[
R_r=\sum_{i=0}^{N-1} \mathbf{1}\{\text{agent alive at macrostep } i\}
\]
Standard CartPole is considered solved when an agent consistently achieves a return of $\approx 500$. In our experiments, depending on the setting, we cap the maximum number of environment steps (and therefore the maximum possible return) at either 500 or 800. An agent that selects actions uniformly at random achieves an average return of $\approx 20-25$.

\subsection{GoL-EM}
\label{sec:exp-gol-cp}

We test Conway's Game of Life (GoL) as a fixed rule cellular-automaton substrate for CartPole control. The question is whether nontrivial control behavior can be obtained by training only the initial GoL state with a fixed encoding/decoding mechanism. In EM notation, the controller has the form:
\[
F_i
=
D\!\left(
f_{\mathrm{GoL}}^{T}
\left(
p\oplus E(x_i)
\right)
\right)
\]
where \(x_i=(q_i,\dot q_i,\theta_i,\dot\theta_i)\) is the CartPole observation at macrostep \(i\), \(p\) is the learned GoL program, \(T\) is the internal GoL computation time budget, and \(F_i\) is the decoded force applied to the cart. In our setting, each episode was capped at \(500\) CartPole macrosteps.

This EM is purely reactive and does not use state retention. At every CartPole macrostep, the board is reset to the same learned program \(p\), the current observation is encoded, the GoL dynamics are run, and the final configuration is decoded into an action.

The learned program \(p\) is placed in an upper-left region of the board. Program is constructed by combining a small number of gliders together with sparse binary perturbations. Interface geometry is fixed: the CartPole variables are written into a designated input region, and the force is read from designated output-collector regions.

\begin{figure}[H]
    \centering
    \includegraphics[width=1\linewidth]{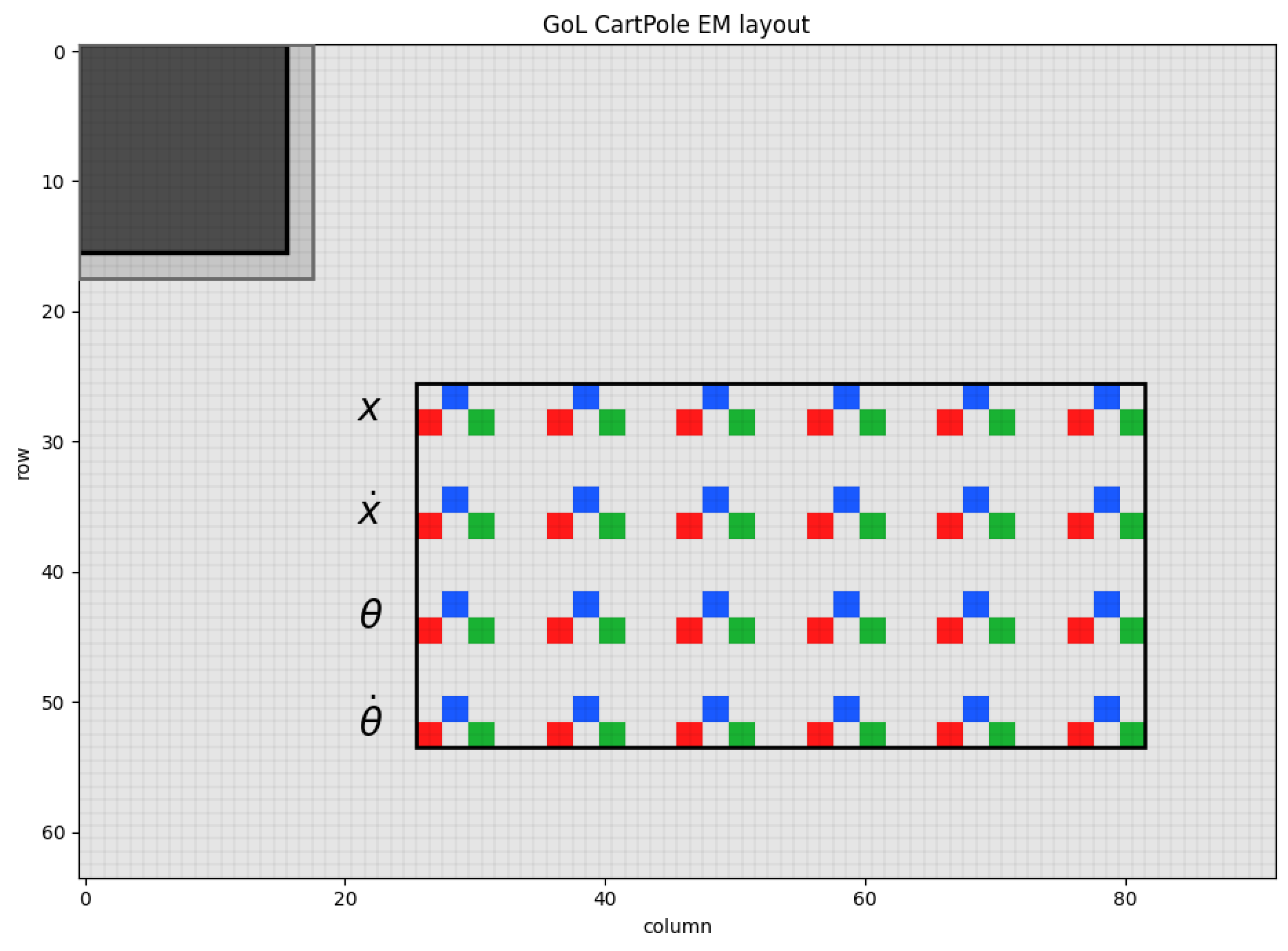}
    \caption{Board layout of the GoL-EM controller for CartPole. The learned GoL program is placed in the upper-left region, separated from the input-output interface by a buffer zone. Four horizontal lanes encode the CartPole variables \(q,\dot q,\theta,\dot\theta\). In each lane, one of six spatial bins is activated according to the discretized value of the corresponding variable. Blue cells indicate input-block locations, while red and green cells indicate left and right output-collector regions used for force decoding.}
    \label{fig:gol-cp-layout}
\end{figure}

The encoder \(E\) maps the continuous CartPole observation to a GoL pattern. Each of the four observation variables is first clamped to a useful range and discretized into one of six bins, ordered from left to right by increasing value. Each variable has its own horizontal lane, and the selected bin in that lane is activated by writing a \(2\times2\) block of live cells. Thus each CartPole observation is represented by four active input blocks, one per variable (see \cref{fig:gol-cp-layout}).

\begin{figure}[H]
    \centering
    \includegraphics[width=0.3\linewidth]{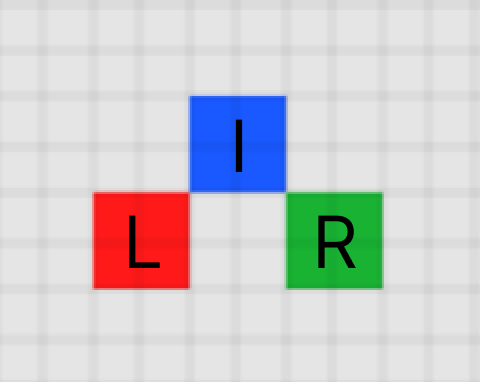}
    \caption{Single-bin motif used in the GoL-EM CartPole interface. If this bin is selected, the blue \(I\) zone is activated as a \(\mathbf{1}_{2\times2}\) alive block. After GoL evolution, live cells in the red \(L\) and green \(R\) regions contribute to the aggregate left and right force votes.}
    \label{fig:gol-cp-bin}
\end{figure}

Each input bin (I) is paired with two nearby output-collector regions, one contributing to a left-force vote (L) and one contributing to a right-force (R) vote. The local motif is shown in \Cref{fig:gol-cp-bin}.

After encoding, the board evolves under the standard Conway's GoL rule for a fixed computation time \(T\). If the configuration reaches a fixed point before this budget is exhausted, i.e. \(s_t=f(s_t)\), the run stops early and we set \(H(s_t)=1\), since further GoL steps would leave the board unchanged. If no stable configuration is reached, the terminal state is still decoded, but a small fitness penalty is applied to encourage halting.

Then, the decoder aggregates activity over all left and right output-collector bins. Let \(L(s_T)\) and \(R(s_T)\) denote the total count of live cells in the left and right collectors at the terminal state \(s_T\). The applied continuous force is:
\[
F
=
\mathrm{clip}
\left(
g_F \frac{R(s_T)-L(s_T)}{R(s_T)+L(s_T)+\varepsilon},
-F_{\max},
F_{\max}
\right)
\]
Where $g_F$ is a force gain hyperparameter, not trained. Thus the direction and magnitude of the force are determined by the imbalance between activity in right and left collectors.

Training uses a mutation-only genetic algorithm over gliders position, with occasional sparse binary mutation on the program region. For one episode, the fitness is \(F=R-\lambda_{\mathrm{nohalt}}\,\rho_{\mathrm{nohalt}}\) , where \(R\) is the CartPole return, \(\rho_{\mathrm{nohalt}}\) is the fraction of CartPole macrosteps in which the GoL computation failed to halt within the time budget, and \(\lambda_{\mathrm{nohalt}}\) controls the strength of this penalty.

Importantly, the interface layout and computation time budget should be understood as simple heuristic design choices, not as the result of automated optimization over many possible interfaces. The budget \(T\) was chosen to allow gliders have enough time to reach the furthest input/output locations and interact, while bin and lane spacing were chosen to limit excessive interference between neighboring regions. More effective interface choices may exist.

\subsubsection{Results}

\begin{figure}[H]
    \centering
    \includegraphics[width=1\linewidth]{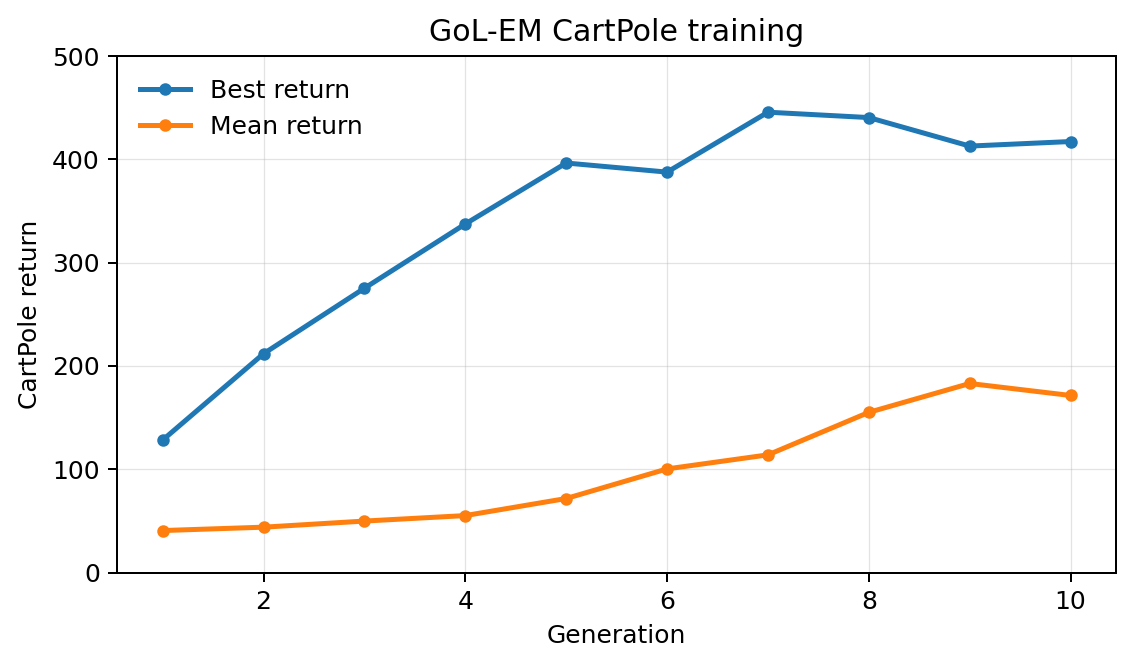}
    \caption{Learning curves for the representative run, showing mean and best CartPole return across generations. The maximum attainable return is 500.}
    \label{fig:gol-cp-curve}
\end{figure}

GoL-EM achieved CartPole returns clearly above the random baseline of $\approx 20$. In the representative training run reported here, the evolved population reached a mean return of \(171.36\), while the best controller reached \(417.20\). We then saved this best controller and evaluated it on \(20\) new episodes, obtaining a return of \(316.90 \pm 121.52\). During evaluation, the mean non-halting rate was \(54.5\%\), meaning that in more than half of the CartPole macrosteps the model didn't halt before maximum time budget. (Results are summarized in \cref{tab:gol_cp_results}, and the learning curve in \cref{fig:gol-cp-curve}).

\begin{table}[H]
\centering
\footnotesize
\setlength{\tabcolsep}{7pt}
\begin{tabular}{@{}lc@{}}
\toprule
Metric & Value \\
\midrule
GoL computation budget \(T\) & 300\\
Population size & 100 \\
Training generations & 10 \\
Final population mean return & 171.36 \\
Best training return & 417.20 \\
Evaluation return & \(316.90 \pm 121.52\) \\
non-halting rate& 54.5\% \\
\bottomrule
\end{tabular}
\caption{Results from the representative run.}
\label{tab:gol_cp_results}
\end{table}

Qualitatively, the learned policy was simple but meaningful. The controller typically adopted a one-sided balancing strategy: it pushed the pole towards one side and then maintained balance through slow cart motion until eventually exiting the allowed position bounds. Since the vast majority of inspected episodes ended by exceeding the cart-position range rather than the pole-angle range, the policy appears to use angular information far more effectively than positional information, possibly ignoring positional information at all.
This suggests that GoL-EM is capable of control behavior, but with several robustness-related limitations.

\begin{figure}[H]
    \centering
    \includegraphics[width=1\linewidth]{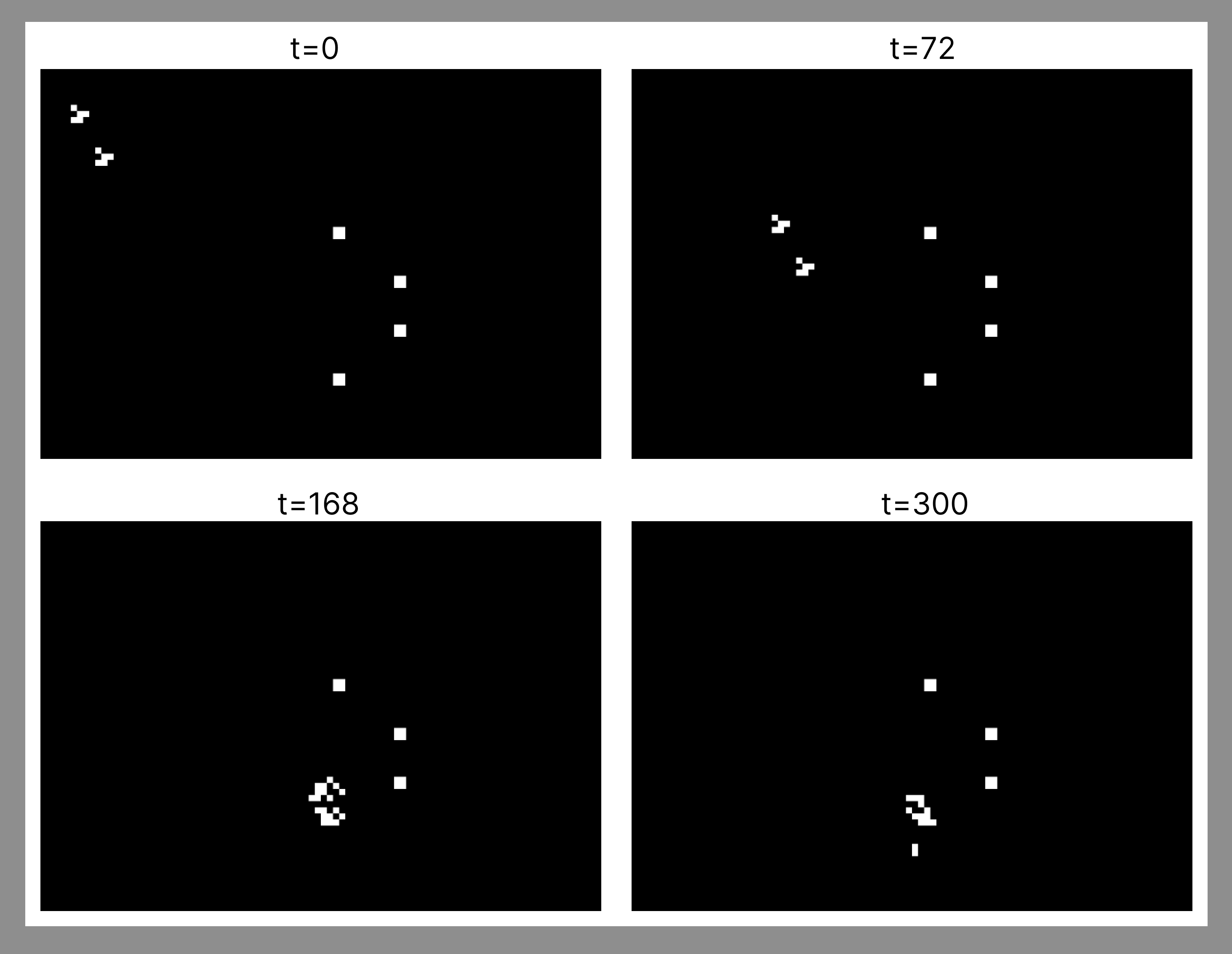}
    \caption{Snapshots of GoL evolution during 1 CartPole macrostep.}
    \label{fig:gol-cp-frames}
\end{figure}

Figure~\ref{fig:gol-cp-frames} shows some representative snapshots of GoL state during a prediction. The dynamics are driven by glider motion and collisions with the encoded input blocks.

We also briefly tested state retention by using the final GoL configuration from one CartPole macrostep to initialize the next (we employed identity retention \(R(s)=s\) ). In the present setting, this made the controller unstable and dropped its performance close to random, as residual debris from previous macrosteps accumulated and interfered with the next encoded observation in near-chaotic ways.

\subsection{CEM1D}
\label{sec:exp-continuous-1d-cp}

We then explored a continuous-valued, one-dimensional Emergent Model (CEM1D) for CartPole. This experiment differs from the previous GoL controller in two ways. First, the substrate is not a discrete binary cellular automaton, but a continuous-valued local dynamical system. Second, the controller uses retention across CartPole macrosteps: its internal state is retained from one control step to the next, rather than being reset after every action.

The latent state is a one-dimensional lattice of $L$ cells, \(s\in\mathbb R^L\) .
Its dynamics are defined by a local radius-1 update rule. In continuous-time form, the rule can be written as:
\[
\frac{ds_j}{dt}
=
a_l s_{j-1}
+
a_c s_j
+
a_r s_{j+1}
-
\gamma s_j^3,
\]
where \(s_j\) is the value of the cell at lattice position \(j\). The coefficients \(a_l,a_c,a_r\) are trainable local interaction weights for the left, center, and right cells, respectively. The cubic term provides nonlinear damping and helps prevent exponential growth of the state. Its coefficient is fixed to \(0.03\), and not trainable.

In our implementation, we integrate with explicit Euler steps of fixed size \(\Delta t=0.1\):
\[
s_j^{t+1}
=
s_j^t
+
\Delta t
\left(
a_l s_{j-1}^t
+
a_c s_j^t
+
a_r s_{j+1}^t
-
\gamma (s_j^t)^3
\right)
\]
Thus the model is implemented as a discrete dynamical system, although being motivated by a continuous-time update rule. Boundary conditions can be periodic, reflective, or zero-padded. In the representative run reported here, we used reflective boundaries.

We fix, once for all, a set of ports into the lattice as positions for writing/reading inputs and outputs.
At each CartPole macrostep, the physical state \(x=(q,\dot q,\theta,\dot\theta)\) is normalized and injected into four input ports by overwrite. If \(x_k\) is the \(k\)-th input component and \(j_k\) is its corresponding fixed port location, the encoding operator is:
\[
E:\qquad s_{j_k}\leftarrow x_k
\]
The system then evolves for a fixed duration of \(T\) microsteps, without employing a halting operator.

After time evolution, a designated output cell \(s_{j_{\mathrm{out}}}\) is read and thresholded to produce the discrete CartPole force:
\[
F_i =
\begin{cases}
-F_{\max}, & s_{j_{\mathrm{out}}} < -\eta,\\
+F_{\max}, & s_{j_{\mathrm{out}}} > \eta,\\
\text{random choice in }\{-F_{\max},+F_{\max}\}, & |s_{j_{\mathrm{out}}}|\le \eta
\end{cases}
\]
Equivalently,
\[
D:\qquad F_i=\operatorname{threshold}(s_{j_{\mathrm{out}}})
\]

The random tie-breaking region around zero is employed to avoid a degenerate control strategy observed in preliminary runs, where the controller exploited extremely small output values, for example on the order of \(10^{-6}\). Such near-zero decisions can be numerically fragile and are also poorly suited to long-term state retention. The chosen decoder therefore enforces robust binary actions.

Only the local interaction parameters \((a_l,a_c,a_r)\) are explicitly trained. The latent state \(s\) is implicitly optimized with Lamarckian-style inheritance (see Section~\ref{sec:experiments}).

\begin{figure}[H]
    \centering
    \includegraphics[width=1\linewidth]{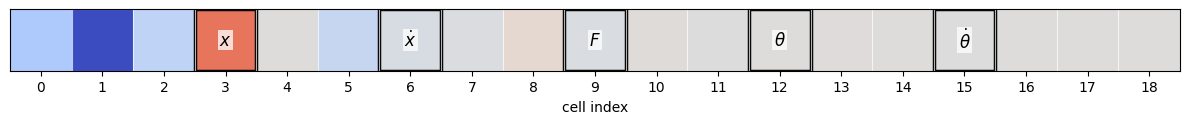}
    \caption{Layout of CEM1D. The CartPole state variables are written into designated input ports, and the force is decoded from the central output port.}
    \label{fig:continuous-1d-cp-interface}
\end{figure}

As a representative minimally engineered configuration, we used a one-dimensional lattice of length \(L=19\). The four CartPole observations are written into input cells at positions \((3,6,12,15)\), and the action is read from a single output cell at the central position \(9\); this places two input cells on each side of the central readout, a deliberately simple symmetric arrangement (\Cref{fig:continuous-1d-cp-interface}). The dynamics are integrated with explicit Euler steps of size \(\Delta t=0.1\) for a fixed computation time of \(T=30\) microsteps per macrostep, with reflecting boundary conditions.

\subsubsection{Results}

In our experiments, the maximum episode length is \(800\) macrosteps. Fitness \(F\) is computed by evaluating the controller over multiple episodes (rollouts) and combining their returns into a single score that rewards high mean return while penalizing variability across rollouts:
\[
F
=
\frac{\operatorname{Mean}_{r}\!\left(R_r\right)}
{1+\lambda_{\mathrm{std}}\operatorname{Std}_{r}\!\left(R_r\right)}
\]
where \(R_r\) is the return obtained in rollout \(r\), the mean and standard deviation are taken over rollouts, and \(\lambda_{\mathrm{std}}\) controls how strongly variability is penalized. Dividing the mean return by an increasing function of its standard deviation favors controllers that perform robustly across different environment initializations, rather than ones that occasionally survive for a long time.

\begin{figure}[H]
    \centering
    \includegraphics[width=1\linewidth]{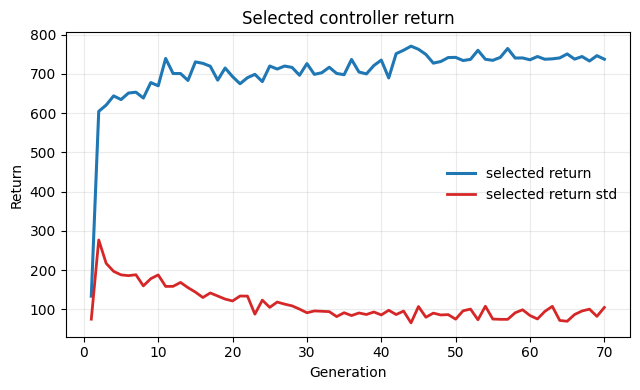}
    \caption{Best controller return and its standard deviation across training generations.}
    \label{fig:continuous-1d-cp-return-curve}
\end{figure}

CEM1D learned high-performing CartPole policies, reaching returns close to the maximum of \(800\) and far above the random baseline of \(\approx 20\). The best controller by training fitness achieved a return of \(737.32 \pm 104.63\) during training. Re-evaluated over \(1000\) new rollouts, it achieved a mean return of \(734.93 \pm 107.00\), confirming that the learned behavior reflects a real control policy rather than a lucky selection bias. (Results are summarized in \cref{tab:continuous_1d_cp_results} and the learning curves in \cref{fig:continuous-1d-cp-return-curve}).

\begin{table}[H]
\centering
\footnotesize
\setlength{\tabcolsep}{6pt}
\begin{tabular}{@{}lc@{}}
\toprule
metric & value \\
\midrule
generations & 70\\
population size & 120 \\
lattice length \(L\) & 19 \\
internal steps \(T\) & 30 \\
input cells & \((3,6,12,15)\) \\
output cell & \(9\) \\
training return & \(737.32 \pm 104.63\) \\
evaluation return & \(734.93 \pm 107.00\) \\
\bottomrule
\end{tabular}
\caption{Results for a representative CEM1D CartPole training run.}
\label{tab:continuous_1d_cp_results}
\end{table}

For the selected controller, the learned update equation is:

\[
\resizebox{\columnwidth}{!}{$\displaystyle
s_j^{t+1}=s_j^t+0.1\left(0.796\,s_{j-1}^t-0.319\,s_j^t-1.628\,s_{j+1}^t-0.03\,(s_j^t)^3\right)
$}
\]

\begin{figure}[H]
    \centering
    \includegraphics[width=0.75\linewidth]{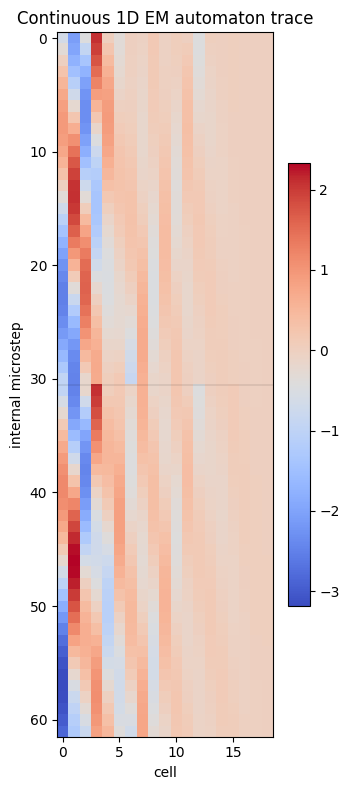}
    \caption{Internal "reasoning" trace of the CEM1D controller during two environment steps. Rows correspond to time evolution in microsteps, and columns correspond to lattice cells in the one-dimensional medium. The horizontal line at $t=31$ marks the boundary between consecutive macrosteps, where the input is re-injected.}
    \label{fig:continuous-1d-cp-trace}
\end{figure}

A representative space-time diagram of the automaton is reported in \cref{fig:continuous-1d-cp-trace}.

The learned policy substantially outperforms the GoL-EM controller and is very robust to deviations in the pole angle, but it still exhibits some fragilities. In nearly all manually inspected rollouts, the cart eventually drifted out of the allowed position range once enough time had passed. This occurred despite using a boundary-heavy initialization distribution, with \(70\%\) of training episodes starting near the cart-position boundaries to encourage robustness. Qualitatively, this suggests that the controller learned an effective pole-balancing strategy, but did not integrate correctly the cart-position information into its decision process.

This limitation is consistent with the structure of the learned dynamics. The update rule is predominantly linear, and this may limit expressivity: the cubic term has a very small coefficient (\(0.03\)) and contributes substantially only when cell values become large, so for most of the operating range the lattice evolves close to a linear system. Two further design choices restrict expressivity: the fixed input and output port layout may constrain how the local dynamics can route information through the lattice, and the fixed computation time \(T\), rather than an adaptive halting mechanism, caps the number of internal steps available per decision.

\begin{figure}[H]
    \centering
    \includegraphics[width=1\linewidth]{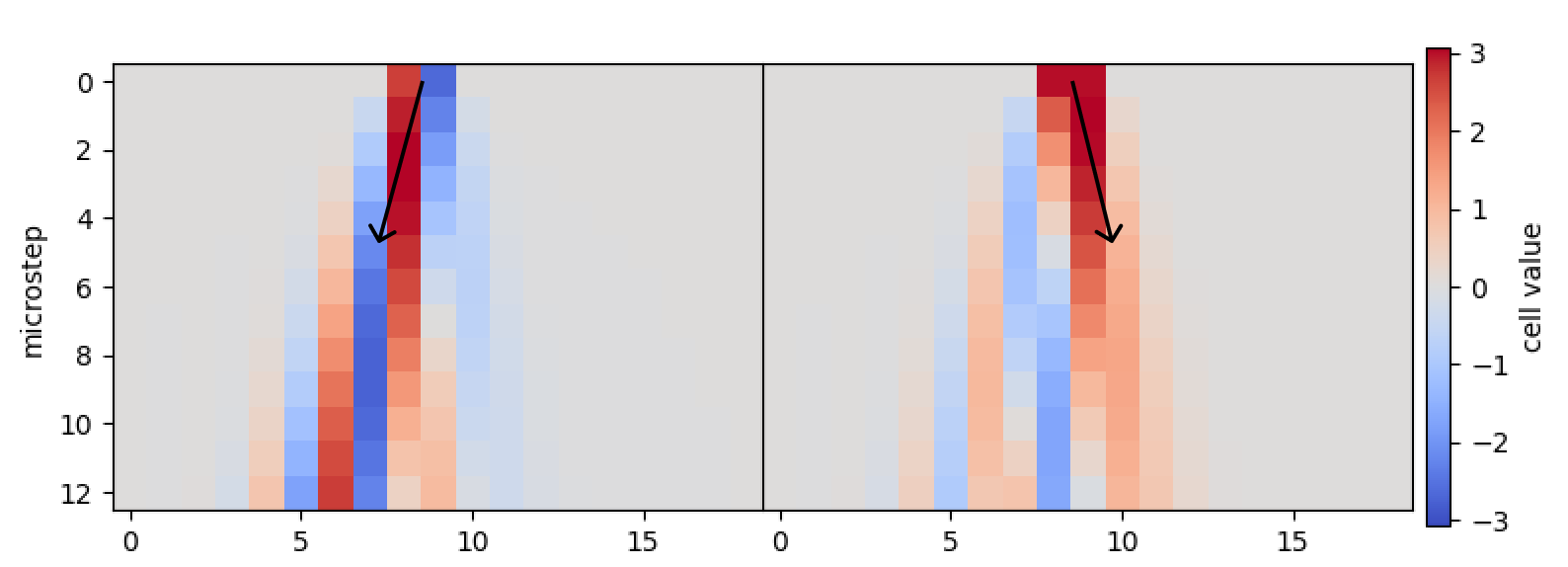}
    \caption{Information transport with right and left propagating patterns.}
    \label{fig:continuous-1d-cp-gliders}
\end{figure}
Interestingly, the evolved dynamics exhibit localized structures that behave like information carriers, similarly to gliders in discrete cellular automata (See \cref{fig:continuous-1d-cp-gliders}). In particular, two nearby cells with similar values tend to propagate their state to the right over successive microsteps, while two nearby cells with opposite values, forming a strong local gradient, tend to propagate information to the left. The left-moving transport appears stronger and more robust than the right-moving. This asymmetry may partly explain why position information, which would need to propagate rightward toward the output port, is used less effectively by the controller.

The CEM1D CartPole experiment should therefore be interpreted as an exploratory feasibility result rather than a mature control method. Nevertheless, the controller achieves high returns despite its limited use of cart-position information, comfortably exceeding the standard solution threshold of \(500\).

\subsection{Meta-Life}
\label{sec:exp-metalife}

\begin{figure}[H]
    \centering
    \includegraphics[width=1\linewidth]{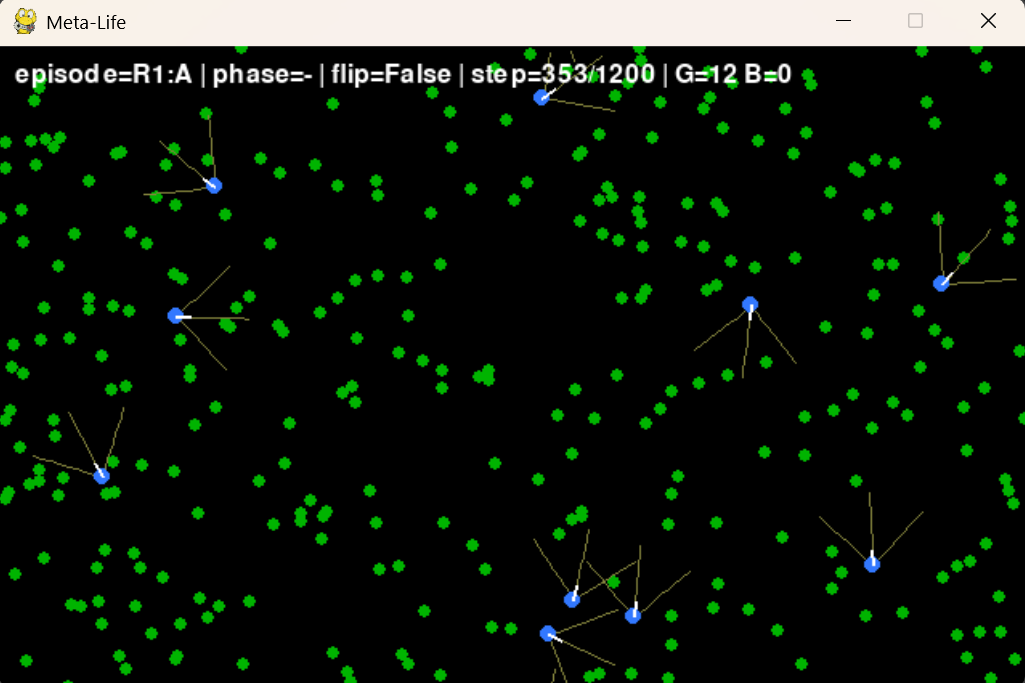}
    \caption{Meta-Life environment. Agents (blue dots) move and collect resources (green dots).}
    \label{fig:metalife-env}
\end{figure}

We next tested Emergent Models in a simple embodied environment, which we call Meta-Life. Meta-Life is a continuous two-dimensional toroidal world in which agents sense, navigate and collect resources (see \cref{fig:metalife-env}).

\begin{figure}[H]
    \centering
    \includegraphics[width=0.5\linewidth]{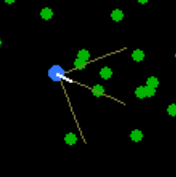}
    \caption{Meta-Life agent with its three lidar rays.}
    \label{fig:metalife-agent}
\end{figure}

Each agent has a position in the 2D world and a heading that defines the direction it faces. It senses the environment through three lidar rays cast at fixed angles relative to its heading, so the rays rotate with the agent. Each ray returns a scalar in \([0,1]\), indicating the proximity of the nearest resource along its direction: \(0\) means no resource is detected, while larger values indicate a closer resource (see \cref{fig:metalife-agent}).

The observation is therefore:
\[
x_i=(\ell_{1,i},\ell_{2,i},\ell_{3,i})
\]
where \(\ell_{1,i},\ell_{2,i},\ell_{3,i}\in[0,1]\) are the lidar signals at macrostep \(i\). The controller outputs the rotational speed \(\omega_i\) and the forward speed \(v_i\), allowing the agent to steer and move.
\[
y_i=(\omega_i,v_i)
\]

We consider two regimes. In \emph{Meta-Life-Food}, every resource is food, and collecting one gives reward \(+1\). This is an ordinary foraging task: the agent must use its lidar observations to steer toward food and collect it.

In \emph{Meta-Life-Adapt}, resources alternate between food and poison in phases of random duration, while remaining visually identical: during a food phase, collecting any resource gives reward \(+1\); the world then switches to a poison phase, in which any resource gives reward \(-1\); it later switches back, and so on. Since food and poison produce identical lidar readings, the agent's observation carry no information about the current phase, and cannot tell in advance whether the next contact will be good or bad. It must therefore periodically "taste" a resource, observe the resulting reward (observations are augmented with a reward signal; see below), and adapt online, switching between a "chase resources" and an "avoid resources" mode without retraining.

Each rollout consists of two mirrored episodes, A and B, each composed of four alternating phases. Episode A runs:
\[
\text{food }(T_1)\to
\text{poison }(T_2)\to
\text{food }(T_3)\to
\text{poison }(T_4)
\]
and Episode B runs the same phases with reversed reward signs:
\[
\text{poison }(T_1)\to
\text{food }(T_2)\to
\text{poison }(T_3)\to
\text{food }(T_4)
\]

Here \(T_1,\dots,T_4\) are phase lengths in macrosteps, resampled independently for each rollout from a uniform distribution over roughly \((0,600)\).

We use two variance-reduction measures. First, the A/B symmetry above makes the score less sensitive to any particular phase-length sample, since each phase ordering is evaluated under both sign assignments. Second, each agent is evaluated over several rollouts per generation (\(N_{\mathrm{rollouts}}\)) and its score averaged. Both limit the influence of a lucky or unlucky sample on the estimated fitness.

The observation is augmented with a transient reward-feedback signal:
\[
x_i=(\ell_{1,i},\ell_{2,i},\ell_{3,i},r_i)
\]
where \(r_i\in\{-1,0,+1\}\). The signal defaults to \(r_i=0\) when no contact occurs; after contact with a resource it is set to the collected reward (\(+1\) for food, \(-1\) for poison) and held for \(4\) macrosteps before returning to \(0\).

For diagnostic purposes, we classify scores by their timing relative to the previous contact. A reward earned from a new contact within \(6\) macrosteps of the previous one falls in the \emph{reaction-window}: the period in which the reward signal from that previous contact may still be present in the controller's input. Rewards earned later fall in the \emph{after-reaction} bucket. 

After-reaction scores are therefore the stricter diagnostic: they measure whether the agent has inferred the current phase and retained it as a persistent internal memory, rather than merely reacting to the instantaneous reward signal. Since a phase lasts on average \(300\) macrosteps, far longer than the \(6\)-macrostep reaction window, good adaptive behavior requires holding this phase memory across many reward-free macrosteps.

For a single agent and rollout, return is the cumulative reward:
\[
R = G-B
\]
where \(G\) is the number of good contacts (collected food) and \(B\) is the number of bad contacts (collected poison).

For Meta-Life-Adapt, we also calculate the meta score:
\[
\mathrm{meta}
=
\frac{G-B}{G+B+\epsilon},
\]
A meta score of \(0\) indicates phase-insensitive behavior, i.e. no adaptation to the changing phases. A score of \(1\) indicates perfect selectivity, with only food contacts and no poison contacts, while a score of \(-1\) indicates the worst case, with only poison contacts and no food contacts. The term \(\epsilon\) is a small constant that prevents division by zero when no contacts occur.
We also report the same score restricted to reaction-window and after-reaction to verify whether adaptation is persistent.

During training, the food-only objective is the raw return reduced by two penalty terms:
\[
F = G - \lambda_{\mathrm{loop}}L - \lambda_{\mathrm{invalid}}I,
\]
where \(G\) is the number of food items collected and \(\lambda_{\mathrm{loop}},\lambda_{\mathrm{invalid}}\) are penalty coefficients. The first term penalizes persistent rotation: \(L\) is a moving average of the agent's angular velocity, so steering sustained in one direction accumulates penalty while brief turns in opposite directions average out. This discourages a degenerate attractor, observed in preliminary runs, in which the agent spins endlessly in place instead of exploring the environment. The term \(I\) is the fraction of macrosteps in which the run becomes numerically invalid, i.e.\ a \texttt{NaN} or \texttt{Inf} appears in the action or in the model's internal state.

Meta-Life-Adapt employs a slightly different objective:
\[
F
=
(G+\epsilon)^{0.5}
-
(B+\epsilon)^{0.5}
-
\lambda_{\mathrm{loop}}L
-
\lambda_{\mathrm{invalid}}I
\]
The square root exponent forces diminishing returns in the absolute number of contacts. This reduces pressure to collect as many resources as possible and instead favors policies that are more selective, i.e. that maintain a good food/poison ratio even if collecting fewer total resources.

\emph{Meta-Life-Adapt} tests in-distribution adaptation, a weaker regime than meta-learning: the agent must switch its policy online from the observed reward, but both regimes are seen during training and the model is never asked to adapt to a novel task. Online policy switching from a feedback signal is nonetheless a precondition for meta-learning, since a system unable to switch policy in response to reward could not adapt to unseen tasks. We therefore view this setting as a precursor to meta-learning.

\paragraph{CEM2D}
CEM2D is a continuous two-dimensional Emergent Model featuring a fully local update rule. Its state has two channels,
\[
s=(c,m),
\qquad
c,m\in\mathbb R^{H\times W}
\]
where \(c\) is a fast activation field and \(m\) a slow memory field; in the experiments below the lattice is square, \(H=W=8\), with each cell indexed by \((i,j)\). 

This state is paired with an interface \((E,D)\) that writes inputs to and reads outputs from specific lattice cells (ports). CEM2D uses no explicit halting predicate \(H\): instead, the reasoning-time budget is itself a trainable scalar \(T\), constrained to the interval \([2,10]\) and held fixed during inference.

The encoder writes each observation component \(x_k\) into a port, a single lattice cell, rather than spreading it densely across the lattice. Each port has a trainable location: the coordinates \((i_k,j_k)\) of its cell are themselves learned parameters, so training decides where each input component enters the lattice. At its ports, the encoder applies a pointwise affine map to the observation and then overwrites the activation state:
\[
c[i_k,j_k] \leftarrow w_k x_k+b_k,
\qquad k\in\{1,2,3,4\},
\]
where \(w_k\) and \(b_k\) are per-input trainable scalars setting the scale and the bias of the \(k\)-th component. The encoder thus commits only a few cells to input, while leaving two things to training: the port locations and their input scaling. This keeps the interface low-dimensional yet flexible.

We employ a fully local update rule \(f\): every cell updates from quantities computed over its own neighborhood. Define the Moore-neighborhood mean of the activation field around cell \((i,j)\):
\[
\sigma_{ij}
=
\frac{1}{8}
\sum_{(u,v)\in N_8(i,j)}
c_{uv}
\]
At each microstep, every cell first computes a scalar preactivation \(q_{ij}\):
\[
q_{ij}
=
a_1
+
a_2 c_{ij}
+
a_3 \sigma_{ij}
+
a_4 m_{ij}
+
a_5 c_{ij}\sigma_{ij}
\]
The activation field is then updated by a residual step:
\[
c_{ij}^{t+1}
=
c_{ij}^t
+
\alpha
\left(
a_0+\tanh(q_{ij}^t)-\lambda c_{ij}^t
\right)
\]
The memory field is updated through a local write gate \(g_{ij}\in[0,1]\), computed from the same preactivation:
\[
g_{ij}^t
=
\operatorname{sigmoid}(g_b+g_k q_{ij}^t)
\]
\[
m^{t+1}_{ij}
=
(1-\beta g_{ij}^t)m^t_{ij}
+
\beta g_{ij}^tc_{ij}^t
\]
The gate controls how strongly each memory cell moves toward the current activation at the same position: a closed gate leaves the memory unchanged, while an open gate moves it toward \(c_{ij}\).

All coefficients are global scalars, shared by all cells. The rates \(\alpha\) and \(\beta\) scale the update step. The fixed leak \(\lambda=0.25\) damps the activation field and helps prevent unbounded growth of \(c\). The biases \(a_0\) and \(a_1\) act in different places: \(a_0\) in the residual activation update, \(a_1\) in the preactivation \(q_{ij}\). The gate is set by its bias \(g_b\) and gain \(g_k\). The transition function \(f\) is the aggregate one-microstep map these equations define over the whole lattice. 
Wrapping boundary conditions are applied.

After the \(T\) microsteps,
\[
s_{0,i}\xrightarrow{f^T}s_{T,i},
\]
the decoder reads two output ports from the activation field \(c\) and maps their values to the action \((\omega_i,v_i)\). As in the encoder, each output port has a trainable location and a per-output affine map, applied pointwise.

The retention operator used here is identity retention, \(R(s)=s\). This is appropriate because CEM2D employs no explicit halting condition, so the terminal state does not satisfy any halting predicate and can be reused directly. The terminal state therefore becomes the program state for the next macrostep:
\[
p_{i+1}=R(s_{T,i})=s_{T,i}
\]

The fields \(c\) and \(m\) are retained not only across macrosteps but also across generations, using the Lamarckian-style state inheritance protocol described in Section~\ref{sec:experiments}: the selected parent's final \(c\) and \(m\) fields are copied directly to its offspring, while crossover and mutation are applied only to the hard parameters.

\paragraph{Memory-augmented RNN baseline}
We compare CEM2D with a small dense memory-augmented recurrent neural network, denoted mRNN, featuring a fast hidden state \(h_i\in\mathbb R^d\) and a slow memory state \(m_i\in\mathbb R^d\), with \(d=6\). Unlike CEM2D, it employs no explicit gating. It performs one update (microstep) per environment macrostep, corresponding to \(T=1\). That is, it performs no looped internal computation, which places it close to standard recurrent controllers.
The update rule is:
\[
h_{i+1}
=
\tanh
\left(
W_x x_i
+
W_h h_i
+
W_m m_i
+
b_h
\right),
\]
\[
m_{i+1}
=
(1-\beta)m_i+\beta h_{i+1},
\]
and decodes to:
\[
y_i
=
W_{\mathrm{out}}h_{i+1}+b_{\mathrm{out}}.
\]
Unlike CEM2D, this model uses dense recurrent matrices, so each hidden unit can depend on every hidden and memory component; its update rule is therefore nonlocal. Its state \((h,m)\) is retained and inherited under the same Lamarckian convention as CEM2D.

\begin{table}[H]
\centering
\footnotesize
\setlength{\tabcolsep}{6pt}
\begin{tabular}{@{}lccc@{}}
\toprule
model & \makecell{hard params\\size(\(E,D,H,f\))} & \makecell{soft params\\size(\(S\))} & structure \\
\midrule
CEM2D & 35 & 128 & local \(8\times 8\) spatial fields \\
mRNN & 117 & 12 & dense vector recurrence \\
\bottomrule
\end{tabular}
\caption{Meta-Life controller sizes}
\label{tab:metalife_model_sizes}
\end{table}

As shown in Table~\ref{tab:metalife_model_sizes}, CEM2D employs relatively few hard parameters (fixed weights), while using a larger number of soft parameters (latent-state size). The mRNN has the opposite structure: it uses more hard parameters, but a smaller latent state. In CEM2D, the latent-state size is decoupled from the number of hard parameters, since the local update rule is shared across all cells and therefore independent of the lattice size. For the mRNN, by contrast, the two are coupled: the dense recurrent matrices make the number of hard parameters grow quadratically with the hidden-state dimension, \(O(d^2)\). A more systematic evaluation would require testing both architectures across multiple latent-state and hard-parameter scales; here, we consider only an mRNN with \(d=6\) and a CEM2D model with an \(8\times 8\) grid.

\subsection{Training setup}
\label{sec:exp-metalife-training}

Both controllers were trained with the same population-based genetic algorithm. At each generation, agents were evaluated over a fixed number of rollouts and were assigned a fitness score; the next generation was then produced through tournament selection, elitism, sparse crossover, and mutation. Under this scheme, an offspring inherited its hard parameters primarily from a base parent, with a donor parent contributing only a sparse subset of genes through crossover; these hard parameters were then mutated. Soft parameters, instead, were inherited unchanged from the base parent.

The food-only environment used population \(A=100\) and was evolved for \(700\) generations. \emph{Meta-Life-Adapt} used population \(A=300\), evolved for \(510\) generations for CEM2D and \(260\) for the mRNN baseline, since CEM2D required longer to converge. The best agent by training fitness was selected and evaluated on a batch of new rollouts. We report representative runs rather than multi-seed averages: for Meta-Life-Adapt each run takes a few days on a consumer laptop, so multi-seed statistics are left for future work. (Training setup is summarized in \cref{tab:metalife_training_setup}).

\begin{table}[H]
\centering
\footnotesize
\setlength{\tabcolsep}{6pt}
\begin{tabular}{@{}llcc@{}}
\toprule
task & model & population \(A\) & generations \\
\midrule
Food-only & CEM2D & 100 & 700 \\
Food-only & mRNN & 100 & 700 \\
Meta-Life-Adapt & CEM2D & 300 & 510 \\
Meta-Life-Adapt & mRNN & 300 & 260 \\
\bottomrule
\end{tabular}
\caption{Training setup for representative Meta-Life runs}
\label{tab:metalife_training_setup}
\end{table}

\subsection{Results}
\label{sec:exp-metalife-results}

\begin{figure}[H]
    \centering
    \includegraphics[width=1\linewidth]{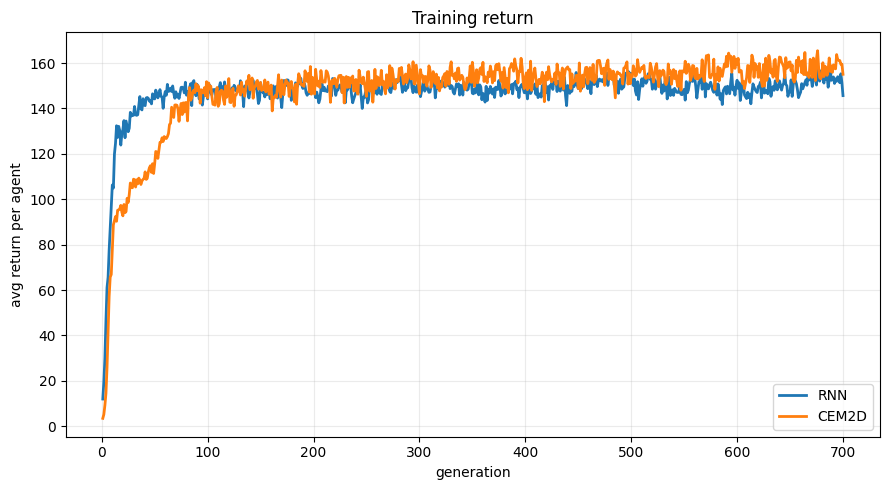}
    \caption{Meta-Life-Food training return per agent per rollout in generations.}
    \label{fig:metalife_foodonly_return}
\end{figure}

In Meta-Life-Food, both controllers learned effective food foraging policies. The mRNN learned faster early in training, while CEM2D caught up later and reached a comparable return. On evaluation, CEM2D reached a return of \(133\) per agent per rollout, while mRNN \(146\).
(Results are shown in \cref{tab:metalife_food_results}, while learning curves in \cref{fig:metalife_foodonly_return}).

\begin{table}[H]
\centering
\footnotesize
\setlength{\tabcolsep}{6pt}
\begin{tabular}{@{}lcc@{}}
\toprule
model & train return & eval return \\
\midrule
CEM2D & 155 & 133 \\
mRNN & 150 & 146 \\
\bottomrule
\end{tabular}
\caption{Meta-Life-Food results. Return is measured per agent per rollout.}
\label{tab:metalife_food_results}
\end{table}

Meta-Life-Adapt was more challenging: the mRNN achieved a meta-score about \(15\%\) higher than CEM2D. Nevertheless, both models obtained high meta-scores relative to the optimum value of \(1\), indicating successful adaptation in both cases. We also observed that CEM2D required substantially longer training and improved more slowly.
(Results are shown in \cref{tab:metalife_adapt_results}).

\begin{table}[H]
\centering
\footnotesize
\setlength{\tabcolsep}{6pt}
\begin{tabular}{@{}lccc@{}}
\toprule
model & meta score & reaction meta & after-reaction meta \\
\midrule
CEM2D & 0.714 & 0.780 & 0.699 \\
mRNN & 0.822 & 0.913 & 0.796 \\
\bottomrule
\end{tabular}
\caption{Meta-Life-Adapt evaluation results. Random baseline is \(0\), perfect behaviour is \(1\).}
\label{tab:metalife_adapt_results}
\end{table}

\begin{figure}[H]
    \centering
    \includegraphics[width=1\linewidth]{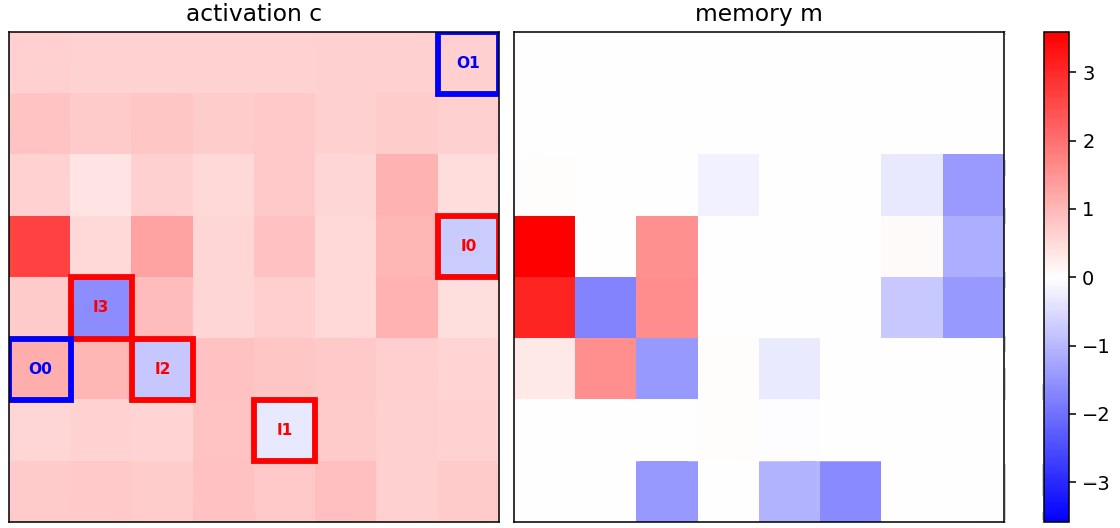}
    \caption{Learned initial condition of the CEM2D controller on Meta-Life-Adapt settings.}
    \label{fig:learned-ic-cem2d}
\end{figure}

Lamarckian inheritance of the \((c,m)\) state shaped the initial condition (program) of the current run, shown in \cref{fig:learned-ic-cem2d}. In the trained model, the coefficients coupling \(c\) to \(m\) are small and the write gates rarely activate, so the memory field \(m\) stays nearly static over time. We then took this trained model, with its inherited state carried across generations, and ablated memory: setting \(m=0\) appears to reduce the meta-score only slightly in the short term, but leads to fully degenerate behavior over longer rollouts.

Two readings fit this: either \(m\) is functionally useful, acting as a static "hardware" that shapes the computation; or \(m\) is inert, and the collapse only reflects that the hard parameters co-adapted to the inherited state during training, so that ablating it moves them off the operating point they were fit to. A conclusive test would require two separate training runs: one with the state inherited across rollouts and generations, as here, and one in which \(m\) is reset to zero at the start of every rollout, retained across macrosteps within a rollout, but never carried across successive rollouts or generations. Comparing the two would show to what extent the model's expressivity comes from the program (inherited state). This is left for future work.

\begin{figure}[H]
    \centering
    \includegraphics[width=1\linewidth]{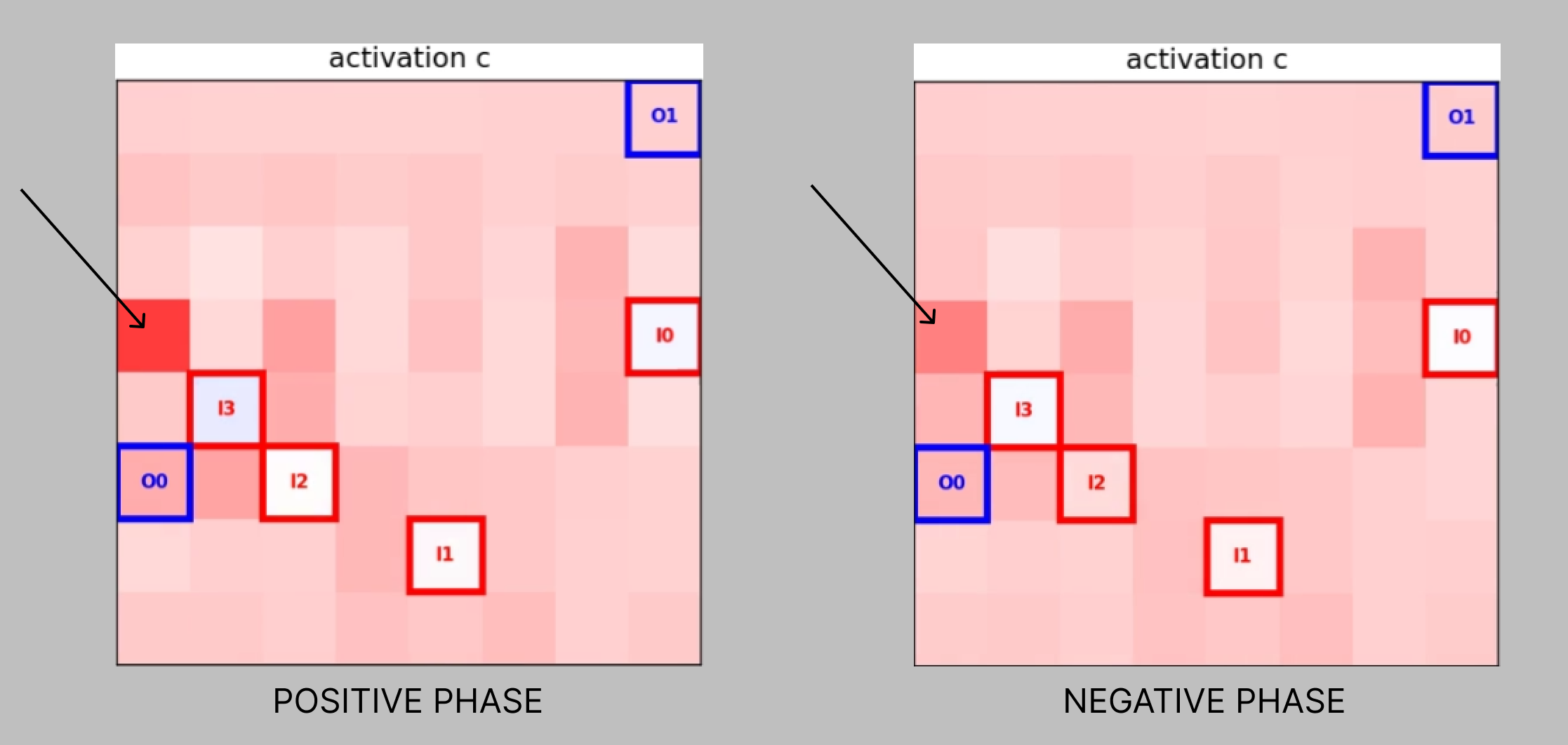}
    \caption{Interpretability visualization of the learned CEM2D dynamics on Meta-Life-Adapt. The two panels show the lattice activation state when the agent estimates a positive (food) phase and a negative (poison) phase. The arrow indicates the regime-correlated cell.}
    \label{fig:cem2d-interpretability}
\end{figure}

Interestingly, during inference we observe a phase-related interpretable feature near the reward input port \(I_3\). The cell in the upper-left neighborhood of \(I_3\) correlates strongly with the agent's estimate of the current resource regime (see \cref{fig:cem2d-interpretability}). Its activation is high after the agent collects food (\(c\approx 3\)) and low after it collects poison (\(c\approx 1\)), tracking the regime the agent has most recently estimated. The cell is highly responsive to reward but has a long effective time constant in its absence, so it updates rapidly on contact yet preserves its value across several reward-free macrosteps.

This cell also appears to modulate behavior. When its activation is high, the agent chases resources and collects them; when it is low, the agent rotates in place, reducing the chance of collecting poison. After the environment switches back to food, an eventual reward reactivates the cell, and the policy returns to resource-seeking. The resulting strategy is simple but effective: seek resources during food phases, rotate in place during poison phases. Although crude, it is sufficient to produce adaptation.

We observe several robustness-related failure modes, most visible on evaluation rollouts substantially longer than those seen during training. The most common is a rotating attractor, present in both CEM2D and in the mRNN: the agent rotates in place indefinitely, especially after a long period in which no resource enters its lidar rays. This occurs despite the looping penalty, though that penalty strongly mitigates it. The mRNN also exhibits a freezing failure: its learned policy permanently sets the velocity to zero after collecting poison, so the agent stops moving entirely and avoids further poison contacts in a risk-averse manner. Both failures share the same consequence: once the agent stops encountering resources, it can no longer taste one, update its regime estimate, or recover, and the policy remains stuck. In CEM2D this is compounded by spontaneous decay of the regime cell, whose activation gradually falls during long stretches without collecting food, pushing the policy toward persistent self-rotation.

Two aspects of the training setup plausibly explain these failures in both models. First, the fitness objective has diminishing returns: the square-root terms reward a high food/poison ratio rather than the raw number of foods collected, so a policy that gathers a few resources and then stops or rotates in place minimizes poison risk while sacrificing little of the expected fitness that further food collection would provide. Second, at each phase change all resources are removed and respawned at random locations. A respawned resource may enter the lidar rays of a rotating agent, or occasionally spawn on top of a frozen one. The agent is therefore rarely required to recover from a stuck state on its own.

\section{Conclusion}
\label{sec:conclusion}

In this work, we introduced Emergent Models as a framework for machine learning through simple, often local, iterated dynamical systems. Rather than treating modeling as the learning of a direct input-output map, EMs treat it as the search for configurations of a computational substrate whose time evolution gives rise to the desired behavior. The ingredients are minimal: a latent state space providing memory, an update rule allowing interactions within that space, an adaptive or sufficiently long computation time, and an interface connecting the substrate to external inputs and outputs.

Beyond abstracting ideas already present in Cellular Automata, Neural Cellular Automata, Neural GPUs, recurrent models, and classical computation theory, the formalism supports a broader position: computation relevant to learning need not reside in a prescribed abstract architecture, such as a neural network, but can emerge directly from a physical-like substrate and its initial conditions. The theoretical ground for this position is the Turing completeness of some simple dynamical systems, such as cellular automata. Classical universality, however, is attributed to the substrate alone: a cellular automaton is universal if some initial configuration simulates a universal Turing machine. Modeling asks for slightly more, namely a fixed, task-independent interface for writing inputs and reading outputs belonging to a general computable domain, such as binary strings \(\{0,1\}^*\), which turns a substrate's ability to simulate arbitrary machines into the ability to compute arbitrary functions as a black box. Latent universality captures this by treating universality as a property of the full Emergent Model \(M = (\mathcal S, f, H, E, D, \oplus)\), comprising substrate, halting condition, and interface, with a strict division of roles: the fixed interface defines only the syntax of communication (how data is written into and read from the substrate), while the program determines the semantics (how the encoded input is interpreted and which computation is performed on it). A minimal example is a universal Turing machine whose tape is split into a program region, a separator, and an input region, accepting inputs written directly as binary strings: program and input are disjoint in the initial condition, so one can search over raw programs without touching the input encoding, i.e.\ the initial state factors as \(p \oplus E(x)\). This division is exactly what a learning setting requires: since the target function is unknown in advance, nothing task-specific can be built into the fixed components, and everything task-specific must live in the learnable program. The latent-universality theorem guarantees that models with this property exist; it does not assert that every substrate universal in the simulative sense is also latent universal.

The experimental contribution is deliberately simpler, intended to validate the base idea of EMs as physical-like learning substrates, aiming to isolate first principles from the effects of scale, engineering, and architectural complexity. Whereas previous local-recursive modeling systems typically employ neural update functions with parameter counts on the order of \(10^5\) or more, the models studied here carry from tens to a few hundred parameters. Within these constraints, a varied set of substrates (discrete cellular automata, Conway's Game of Life, continuous one- and two-dimensional media) exhibits extrapolative computation, closed-loop control, and simple forms of online adaptation (Tables~\ref{tab:model_structure_summary} and~\ref{tab:model_behavior_summary}). The development required to make such systems efficient or competitive remains almost entirely open, and the current state of local-recursive modeling is arguably comparable to that of deep learning two decades ago.

\begin{table}[H]
\centering
\footnotesize
\setlength{\tabcolsep}{4pt}
\begin{tabular}{@{}lllll@{}}
\toprule
model & substrate & hard params & soft params & param domain \\
\midrule
EM43 & 1D CA & 60 & 10 - 50*& \(\{0,1,2,3\}\) \\
GoL-EM (arith.) & 2D GoL & 0 & 100 - 300*& \(\{0,1\}\) \\
GoL-EM (CP) & 2D GoL & 0 & 100 - 300*& \(\{0,1\}\) \\
CEM1D & 1D CEM & 3 & 19 & \(\mathbb R\) \\
CEM2D & 2D CEM & 35 & 128 & mixed\(^{**}\)\\
mRNN & dense RNN & 117 & 12 & \(\mathbb R\) \\
\bottomrule
\end{tabular}
\caption{Structural summary of the experimental models. 
\(^*\) For EM43 and GoL, soft parameters are indicated as program region size. 
\(^{**}\)For CEM2D: real-valued soft and update-rule parameters and integer-valued port locations.}
\label{tab:model_structure_summary}
\end{table}

\begin{table}[H]
\centering
\footnotesize
\setlength{\tabcolsep}{5pt}
\begin{tabular}{@{}ll@{}}
\toprule
model & capabilities \\
\midrule
EM43 & extrapolation on simple arithmetic tasks \\
GoL-EM (arithmetic)& limited capabilities on arithmetic tasks \\
GoL-EM (CartPole)& partially capable of control, not robust\\
CEM1D & stronger control capabilities, still not robust \\
CEM2D & capable of control and adaptation \\
mRNN & capable of control and adaptation, slightly stronger \\
\bottomrule
\end{tabular}
\caption{Behavioral summary of the experimental models.}
\label{tab:model_behavior_summary}
\end{table}

Among the individual results, the clearest is provided by EM43, which learns several simple integer functions and extrapolates perfectly far beyond the training range, including periodic functions that feed-forward neural networks cannot represent globally. In successful runs, generalization is not hidden in an opaque parameter vector: it appears as a stable space-time mechanism in the automaton's diagram, a geometry of interaction that remains valid as the input scale increases. We refer to this phenomenon as \emph{geometric grokking}, and it suggests that time- and translation-invariant substrates can host compact algorithmic routines whose structure is directly visible in their trajectories. Whether geometric grokking persists as task complexity increases, and whether it is partly an artifact of the particular tasks and interface choices employed, remains to be established.

Game of Life was difficult to exploit for learning. Its dynamics are highly sensitive to the initial condition: small changes in the program usually produce large, almost chaotic changes in behavior. This yields a brittle genotype-to-phenotype map and a weakly structured search space, in which similar programs have entirely different fitness, so the search has little local information to exploit and random sampling performs almost as well as evolutionary optimization. The difficulty is compounded by interface design: the interface adopted here, based on isolated \(2\times2\) blocks, may be too poor to reach the computational richness that makes GoL universal, and similarly for the fixed-point halting condition. Even under these constraints, notably, the learned CartPole controller scores substantially above a random policy. Continuous realizations behave differently: naturally suited to control tasks with real-valued inputs and outputs, they are more trainable and less brittle in their genotype-to-phenotype map, though still harder to train than feed-forward or standard recurrent models, which perform a single update per macrostep \(T=1\).

The control experiments are further limited by the simplicity of the environments. CartPole is solvable by linear feedback controllers. In the non-adaptive Meta-Life setting, food attraction and poison avoidance may likewise be solvable by near-linear policies over the sensory inputs; even in adaptive Meta-Life, the required behavior may be achievable by near-linear policies operating on two time scales, fast action selection and slow phase memory, plus a nonlinearity acting as a fast switch that flips the phase memory on regime changes. More complex environments are therefore needed to evaluate the empirical advantages and limitations of Emergent Models relative to conventional neural controllers, especially in terms of generalization, adaptation, and robustness.

A single computational primitive ties the theoretical and empirical parts of this work together: iteration. We conjecture that applying a simple update rule for an adaptive number of steps is what lets a model extrapolate beyond a bounded input range, whereas a single forward pass only interpolates within it. This holds well beyond Emergent Models: adaptive-computation-time RNNs, reasoning transformers, and looped transformers \citep{loopedllm} all iterate, and all extrapolate and generalize better than plain feed-forward networks.

Among iterated systems, we conjecture that a further quantity governs the balance between generalization and trainability: the ratio of parametrization to temporal depth,
\[
q \;=\; \frac{N_{\mathrm{params}}}{\bar T}
\]
where \(N_{\mathrm{params}}\) is the parameter size of the model and \(\bar T\) is the average temporal depth, the mean number of update-function applications per prediction typically employed. In an Emergent Model, \(N_{\mathrm{params}}\) decomposes as program size plus update-rule and interface size, all measured in bits; in a neural network it is the weight memory. Complex behavior can arise at either end: from a large update function applied few times, or from a simple one iterated for long. The ratio admits an Occam's-razor reading, as a form of description length per unit of computation time: low \(q\) forces a short description, iterated for long, to fit the data, pressuring the search toward compressed, rule-like solutions rather than storing data points in memory; high \(q\) grants enough capacity to memorize the data directly, with little pressure to compress and no inductive bias toward algorithmic solutions. Large feed-forward networks are the high-\(q\) limit: all parametrization, no iteration. The conjectured trade-off is that the generalization advantage of low \(q\) is paid in trainability: because the few parameters are reused at every step, a small change in them may compound, and this makes the search landscape irregular. Our evidence for both halves is partial, and whether the trade-off is fundamental, or some substrates can be both well-biased for algorithmic solutions and efficiently searchable, remains open.

The preference for low \(q\) holds where the data is generated by a compressible rule, as in algorithmic, procedural-like tasks. We view such procedural structure as an important component of reasoning: deriving an answer by applying a sequence of operations, rather than merely recalling stored information; indeed, a formal deduction process can itself be viewed, to a large extent, as an algorithm. Memory-intensive tasks are different: factual knowledge is largely incompressible and must be stored somewhere, and no amount of iteration can substitute for storage capacity. This suggests a possible division of competence: high-\(q\) models, such as LLMs, for storing and retrieving factual data; small strongly-recursive models for algorithmic computation, and hence for reasoning-like tasks.
A latent-universal EM, however, could in principle combine both advantages. Its parametrization splits into an update function, whose size is independent of the substrate's spatial extent because it is local, and a program, living in the latent state and variable in size. Memory capacity is added by extending the program, without changing the update function, so the parametrization grows only by the program bits the task requires. The construction is compositional: the same fixed rule serves simple and memory-heavy tasks alike, with a small penalty biasing the program toward simplicity at equal performance while allowing it to grow in length when helps. In this way \(N_{params}\) tracks the intrinsic complexity of the task rather than being an a-priori design choice. A large neural network, by contrast, fuses storage and computation into a single dense map of fixed size, carried in full whether the task requires it or not.

A second motivation is more speculative. The living world realizes adaptive, intelligent behavior even without nervous systems, through simpler physical and chemical mechanisms; and the nervous system itself is likely not fundamental, but a structure that physics supports: neurons obey physical laws and can be regarded as latent structures on top of a lower-level physics. Emergent Models operate at this more basic level, evolving a local rule and an initial configuration rather than a prescribed architecture. This does not exclude neural-network-like organization: if such a structure were the best solution for a task, it could in principle emerge within a latent-universal model. It does, however, reframe a question we cannot yet answer: whether the better route to intelligence on silicon is to build highly simplified models of biological neural networks, as artificial neural networks do, or to construct an artificial physics from which arbitrary structure can emerge. The Bitter Lesson would suggest the latter.

Two concrete directions follow. The first is to realize the latent-universal regime in practice, through a co-evolution of rules and programs under a fixed interface and halting operator. In a first phase, a rule shared across a distribution of tasks is trained jointly with a separate program per task, and fitness selects for rules under which every task admits a solving program, pressuring toward a general rule. In a second phase, that rule is frozen and only the program is trained, on novel tasks of increasing complexity. Success would mean a fixed model that represents new functions by program changes alone. The second direction concerns scale, and bears directly on the trainability of low-\(q\) substrates: progress requires optimization methods built specifically for their nonlocal and nonconvex parameter-to-fitness landscapes, outperforming both pure evolutionary search and gradient-based methods. Such algorithms would apply not only to EMs but also to small recursive neural networks such as TRMs, HRMs, and NCAs.

\section{Acknowledgments}
We thank the Wolfram Institute for research guidance, institutional affiliation, and support.
We thank ResearchHub for supporting this project through an innovative decentralized science funding model.
We are also grateful to everyone who contributed to this work through discussions, ideas,
experiments, and encouragement.
AI tools were used to assist with the implementation of the simulation code and language polishing of the manuscript. Outputs were verified for correctness and consistency. All scientific ideas, interpretations, and conclusions were developed by the authors.

{\footnotesize
\bibliographystyle{plainnat}
\bibliography{references}
}

\appendix
\section{Appendix}

\subsection{Turing machines as Emergent Models}
\label{app:tm-em}

\begin{figure}[H]
    \centering
    \includegraphics[width=1\linewidth]{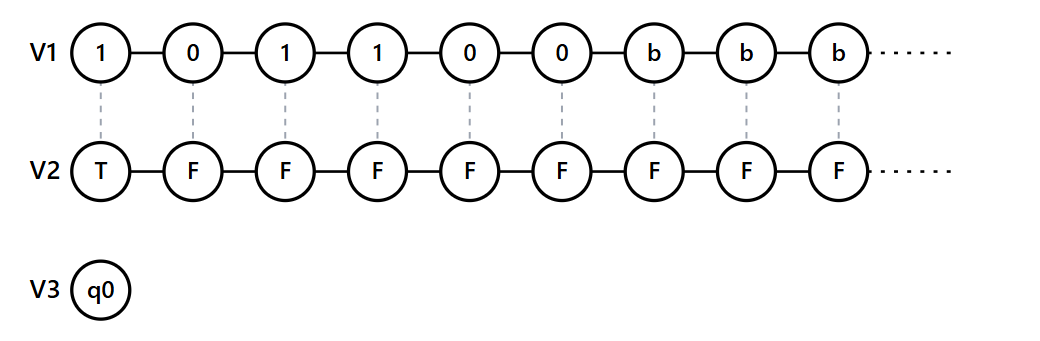}
    \caption{TM generalized automata scheme, with string on tape "101100", head at position 0 and head state q0.}
    \label{fig:automata}
\end{figure}

Turing Machines can be represented as generalized automata (see Figure~\ref{fig:automata} for a graphical representation).

Let \(M=(\Gamma,\delta,Q,q_0,q_{\mathrm{halt}})\) be a Turing machine with tape alphabet \(\Gamma\) containing a blank symbol \(b\), finite control set \(Q\), and transition rule \(\delta\). 
Construct a graph for a generalized automaton made of three parts: a one-sided tape
\(V_1=\{v_i:i\in\mathbb N\}\)
storing actual tape symbols, and, parallelly, a marker tape
\(V_2=\{u_i:i\in\mathbb N\}\)
indicating head position with a unique true marker \(\mathsf{T}\), with all other positions set to false \(\mathsf{F}\), and a single disjoint control vertex \(V_3=\{c\}\) storing the head state in \(Q\). 

The global alphabet is \(W := \Gamma \cup Q \cup \{\mathsf{T},\mathsf{F}\}\)
and the state space is \(\mathcal S := W^{V_1\cup V_2\cup V_3}\)
The halting predicate reads the control vertex and returns \(1\) exactly when \(c=q_{\mathrm{halt}}\).
The global transition \(f:\mathcal S\to\mathcal S\) performs one Turing-machine step via a combination of local and non-local computable operations: locates the unique head position marker in \(V_2\), reads the scanned tape symbol at the corresponding index in \(V_1\), reads the control cell \(c\), applies \(\delta\), writes the new symbol, moves the marker, and updates the control cell. Starting from a well-formed condition, the generalized automaton \((\mathcal S,f,H)\) simulates \(M\) with one microstep per TM step.
The construction is given for a one-sided, single-tape machine, but extends directly to a two-sided tape by taking \(V_1=V_2=\mathbb Z\), and to a \(k\)-tape machine by using \(2k\) tapes in the automaton representation.

\subsection{Universal Turing machines}
\label{app:utm-em}
A universal Turing machine \(U\) simulates an arbitrary Turing machine \(M\) on an input \(x\) when initialized as \(\langle p_M,x\rangle\), where \(p_M=\langle M\rangle\) is a finite program describing \(M\), and \(\langle\cdot,\cdot\rangle\) is a pairing convention that compiles program and input into the initial state of the tape. This convention may also rewrite the raw input into the concrete representation expected by the dynamics of \(U\), for example through block expansions. Crucially, it is a fixed computable initialization scheme for the particular machine \(U\), and does not depend on the machine \(M\) or on the input \(x\) being simulated.

If the combining operator \(\oplus\) is allowed to be an arbitrary computable function \(\mathcal S\times\mathcal S\to\mathcal S\), the traditional UTM notation can be expressed in the EM form \(s_0 = p_M \oplus E(x)\): any pairing scheme \(\langle p_M,x\rangle\) can be rewritten by letting \(E(x)\) apply some transformation to the input and \(\oplus\) perform the arbitrarily complex computable pairing with the program.

We are, however, particularly interested in constructions that admit a disjoint formulation: program and encoded input occupy separate regions, or separate tapes, \(\oplus\) merely places them side by side, and the input encoder has no information about the program nor about the simulated machine \(M\). Examples are the standard textbook three-tape UTM construction or, on a single tape, Watanabe's 5,8 UTM \citep{strieker2020smallutm} and Rendell's SUTM \citep{rendell2016turing}. In the latter two, the tape is spatially divided into two regions: a program region holding the description of the machine to be simulated, and a work/data region holding the encoded simulated tape, therefore an encoding of the input at initialization and of the output at halting.
This makes such constructions easy to interpret as EMs: \(p_M\) stores the program, \(E(x)\) writes the input into the working region, \(D(s)\) reads the output from that region at halting, and \(\oplus\) concatenates/overwrites program and working regions with a delimiter.

\subsection{Proof sketch of Theorem~\ref{thm:lu}}
\label{app:lu-proof}
Let \(g:\{0,1\}^*\rightharpoonup\{0,1\}^*\) be an arbitrary partial computable function. By definition, there exists a Turing machine \(M_g\), such that, for every \(x\in\{0,1\}^*\), when started on tape \(x\,b^\infty\) it halts with tape \(g(x)\,b^\infty\) whenever \(g(x)\) is defined, and does not halt otherwise. Thus \(M_g\) computes \(g\).

Fix a strongly universal Turing machine \(U\) admitting a disjoint program-input formulation (see Appendix~\ref{app:utm-em}), together with its compilation map \(M\mapsto p_M=\langle M\rangle\) and its computable pairing convention \(\langle p_M,x\rangle\), so that \(U\) simulates \(M\) on input \(x\) and halts in a form that preserves the output in a fixed readable format. This choice defines an Emergent Model \((\mathcal S,f,H,E,D,\oplus)\) with fixed substrate and interface: \((\mathcal S,f,H)\) is \(U\) itself, represented as a generalized automaton (Appendix~\ref{app:tm-em}), with \(H\) firing exactly when the head reaches the halting state; the initialization pairing \(\langle p_M,x\rangle\) factors as \(p_M\oplus E(x)\), and the decoder \(D\) reads the output from the tape at halting.

For the target function \(g\), let \(p:=\langle M_g\rangle\). Initialized as \(p\oplus E(x)\), the generalized automaton simulates \(M_g\) on \(x\): if \(g(x)\) is defined, the computation halts and the decoder \(D\) returns \(g(x)\); otherwise the output is undefined. Hence the induced map \(\Phi_p\) computes \(g\) on every \(x\).

Finally, \(p\) and \(p\oplus E(x)\) are finite support (by strong universality), implying that every partial computable binary function \(g\) is realizable by a finite-support initial latent state (finite non-blank input word, followed by an infinite blank background).
\qed

\begin{lemma}[GoL admits latent universality]
\label{lem:gol-lu}
There exists an Emergent Model \(\mathsf M=(\mathcal S,f,H,E,D,\oplus)\) whose automaton is Conway's Game of Life on \(\mathbb Z^2\), such that for every partial computable function \(g:\{0,1\}^*\rightharpoonup\{0,1\}^*\) there exists a finite-support pattern of cells \(p\in\mathcal S\) for which \(\Phi_p\) computes \(g\).
\end{lemma}

\begin{figure}[H]
    \centering
    \includegraphics[width=1\linewidth]{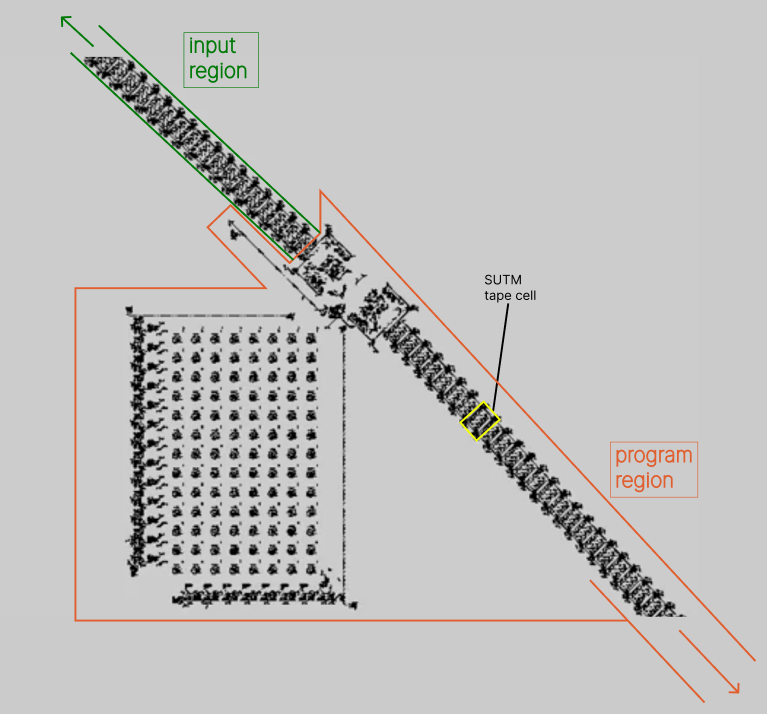}
    \caption{Rendell's UTM in GoL, organized as an Emergent Model with distinct regions. The program is a finite-support initial GoL pattern occupying a disjoint region from the encoded input. The encoder $E(x)$ overwrites on an input region organized in fixed modular patterns corresponding to the UTM tape cells and their symbols.}
    \label{fig:gol}
\end{figure}

\paragraph{Sketch.}
Treat GoL as a generalized automaton with state space \(\mathcal S=\{0,1\}^{\mathbb Z^2}\) and global update \(f\) given by one GoL step. Rendell's construction \citep{rendell2016turing} embeds a universal Turing machine in GoL (Figure~\ref{fig:gol}) together with a fixed compilation convention mapping a pair \((\langle M\rangle,x)\) to an initial state of the GoL grid, with finite support. In that construction, the simulated machine description and the simulated working tape occupy spatially distinct regions, so the initialization can be written in EM form as \(p_M\oplus E(x)\), where \(p_M\) is the finite GoL pattern encoding \( M\), and \(E(x)\) writes the input into the designated data region, and a fixed decoder \(D\) reads the output from the halted configuration. For any partial computable \(g\), let \(M_g\) be a Turing machine computing \(g\), with the corresponding GoL program pattern \(p:=p_{M_g}\). Then the GoL evolution from \(p\oplus E(x)\) simulates \(M_g(x)\), halts if \(g(x)\) is defined, and returns \(g(x)\) after decoding. Hence \(\Phi_p\) computes \(g\). \qed

\subsection{Computation and Modeling on the Continuous}
\label{app:cont-comp}

In the main text, we define exact computation for discrete functions and arbitrary-precision approximation for continuous-valued functions. Here we make the latter notion more explicit.

Standard computable analysis usually treats real numbers through infinite representations (a real input may be represented by an infinite name). A machine can then read as much of this infinite representation as needed to produce an output to a requested accuracy.

We use a different (and simpler) formulation: inputs are discretized to arbitrary precision and encoded as finite strings. This keeps the framework close to ordinary computation theory.

\subsubsection{Finite dyadic representations}

Let
\[
\mathbb{D}
=
\left\{m2^{-q}:m\in\mathbb{Z},\ q\in\mathbb{N}\right\}
\]
be the set of dyadic rationals. These are numbers with finite binary representations, and can therefore be encoded as finite strings.

The set \(\mathbb{D}\) is dense in \(\mathbb{R}\). Hence, for every real input \(x\in\mathbb{R}\) and every input tolerance \(\delta>0\), there exists a finite dyadic approximation \(x_{\mathrm{fin}}\in\mathbb{D}\) such that:
\[
|x-x_{\mathrm{fin}}|<\delta
\]
Equivalently, by increasing the number of binary digits, the dyadic grid becomes finer, and finite dyadic numbers can approximate any real number arbitrarily closely.

Thus, the model receives a finite dyadic approximation of the input, \(x_{\mathrm{fin}}\), whose precision can be increased by using a longer finite string.

\subsubsection{Exact computation on dyadic inputs}
On dyadic inputs, computation is ordinary discrete computation. A model \(\Phi\) parametrized by a finite program \(\theta\) induces a partial map:
\[
\Phi_\theta:\mathbb{D}\rightharpoonup\mathbb{D}
\]

Let:
\[
g_{\mathbb{D}}:\widetilde{\mathbb{D}}\subseteq\mathbb{D}\rightharpoonup\mathbb{D}
\]
be a dyadic-valued partial computable function, total computable on the subset \(\widetilde{\mathbb{D}}\). We say that \(\Phi_\theta\) computes \(g_{\mathbb{D}}\) when:
\[
\Phi_\theta(x_{\mathrm{fin}})
=
g_{\mathbb{D}}(x_{\mathrm{fin}})
\qquad
\text{for all }x_{\mathrm{fin}}\in\widetilde{\mathbb{D}},
\]
and \(\Phi_\theta(x_{\mathrm{fin}})\) is undefined (never halts), for inputs outside \(\widetilde{\mathbb{D}}\).

A universal Turing machine operating on finite dyadic representations can compute exactly any partial computable dyadic function, halting on inputs in the domain and diverging otherwise.

\subsubsection{Continuous computation as arbitrary-precision approximation}
Now let:
\[
g:\widetilde{\mathcal X}\subseteq\mathbb{R}\to\mathbb{R}
\]
be a continuous-valued partial function, total in the domain \(\widetilde{\mathcal X}\).

In this setting, we use the word \emph{computes} in the arbitrary-precision approximation sense, as input and output are finite precision dyadic representations.

For every output tolerance \(\varepsilon>0\), there exists a finite program \(\theta_\varepsilon\) such that, for every real input \(x\in\widetilde{\mathcal X}\), there exists an input tolerance \(\delta>0\) with the following property:
\[
|x-x_{\mathrm{fin}}|<\delta
\quad\Longrightarrow\quad
|\Phi_{\theta_\varepsilon}(x_{\mathrm{fin}})-g(x)|<\varepsilon
\]

The output tolerance \(\varepsilon\) is fixed across the whole domain. The required input tolerance \(\delta\), however, may depend on the particular input \(x\), since some regions of the domain may be more sensitive to input perturbations and therefore require finer approximations. Thus, we do not require a single global input precision that works uniformly for all real inputs. We require only that each real input admits some finite precision level sufficient to achieve the requested output tolerance.

\subsubsection{Asymptotic form}
The same idea can be written asymptotically in terms of input precision. 
We consider the case where, for every real input \(x\), the finite dyadic approximation becomes exact in the limit: the input tolerance tends to zero, \(\delta\to 0\), and therefore \(x_{\mathrm{fin}}\to x\).

Fix an output tolerance \(\varepsilon>0\), and let \(\theta_\varepsilon\) be the corresponding finite program. If the limit exists for every \(x\in\widetilde{\mathcal X}\), define the limiting function \(h\) induced by \(\theta_\varepsilon\) as:
\[
h_{\theta_\varepsilon}(x)
=
\lim_{\substack{x_{\mathrm{fin}}\to x\\ x_{\mathrm{fin}}\in\mathbb{D}}}
\Phi_{\theta_\varepsilon}(x_{\mathrm{fin}})
\qquad
\text{for all }x\in\widetilde{\mathcal X}
\]
The model computes \(g\) if, for every \(\varepsilon>0\), there exists a finite program \(\theta_\varepsilon\), for which this limiting function exists and uniformly approximates \(g\) over the whole domain:
\[
|h_{\theta_\varepsilon}(x)-g(x)|<\varepsilon
\qquad
\text{for all }x\in\widetilde{\mathcal X}
\]
Equivalently,
\[
\left|
\lim_{\substack{x_{\mathrm{fin}}\to x\\ x_{\mathrm{fin}}\in\mathbb{D}}}
\Phi_{\theta_\varepsilon}(x_{\mathrm{fin}})
-
g(x)
\right|
<\varepsilon
\qquad
\text{for all }x\in\widetilde{\mathcal X}
\]

Thus, for each requested output tolerance \(\varepsilon\), there exists a finite program whose limiting behavior on increasingly precise finite inputs stays within \(\varepsilon\) of the target function everywhere on the domain.

A stronger theoretical question, not addressed here, is whether as \(\varepsilon\to0\), the family of programs \(\theta_\varepsilon\) admits a limit, implying that exact asymptotic computation could be realized by a program \(\theta\).

\end{document}